\documentclass[twoside]{article}

\usepackage[accepted]{sty/aistats2022}
\pdfoutput=1

\usepackage[round]{natbib}

\usepackage{algorithm}
\usepackage{algpseudocode}
\usepackage{hyperref}
\usepackage{url}
\usepackage{amsmath,amssymb,amsthm}
\usepackage{amsfonts}
\usepackage{bm}
\usepackage{bbold}
\usepackage[inline]{enumitem}
\usepackage{booktabs}
\usepackage{subfig}

\usepackage{graphicx}
\usepackage{sidecap}
\usepackage{caption}
\usepackage{xspace}

\newcommand*{\ie}{\textit{i.e.}\@\xspace}

\newcommand*{\wrt}{w.r.t.\@\xspace}
\newcommand*{\cfr}{\textit{cfr.}\@\xspace}

\DeclareMathOperator{\vect}{vec}

\DeclareMathOperator{\den}{den}
\DeclareMathOperator{\dev}{dev}

\begin{document}

\runningauthor{Gamba, Chmielewski-Anders, Sullivan, Azizpour,  Bj\"{o}rkman}

\twocolumn[

\aistatstitle{Are All Linear Regions Created Equal?}

\aistatsauthor{Matteo Gamba \And Adrian Chmielewski-Anders \And Josephine Sullivan \And Hossein Azizpour }
\aistatsauthor{M\r{a}rten Bj\"{o}rkman}

\aistatsaddress{KTH Royal Institute of Technology, Stockholm, Sweden} ]

\begin{abstract}
    The number of linear regions has been studied as a proxy of complexity for ReLU networks. However, the empirical success of network compression techniques like pruning and knowledge distillation, suggest that in the overparameterized setting, linear regions density might fail to capture the effective nonlinearity. In this work, we propose an efficient algorithm for discovering linear regions and use it to investigate the effectiveness of density in capturing the nonlinearity of trained VGGs and ResNets on CIFAR-10 and CIFAR-100. We contrast the results with a more principled nonlinearity measure based on function variation, highlighting the shortcomings of linear regions density. Furthermore, interestingly, our measure of nonlinearity clearly correlates with model-wise deep double descent, connecting reduced test error with reduced nonlinearity, and increased local similarity of linear regions.

\end{abstract}

\section{INTRODUCTION}
\label{sec:intro}
    \begin{figure}[t]
	\begin{center}
		\includegraphics[width=0.43\textwidth,trim=1.05cm 1.3cm 1.75cm 2.13cm,clip=true]{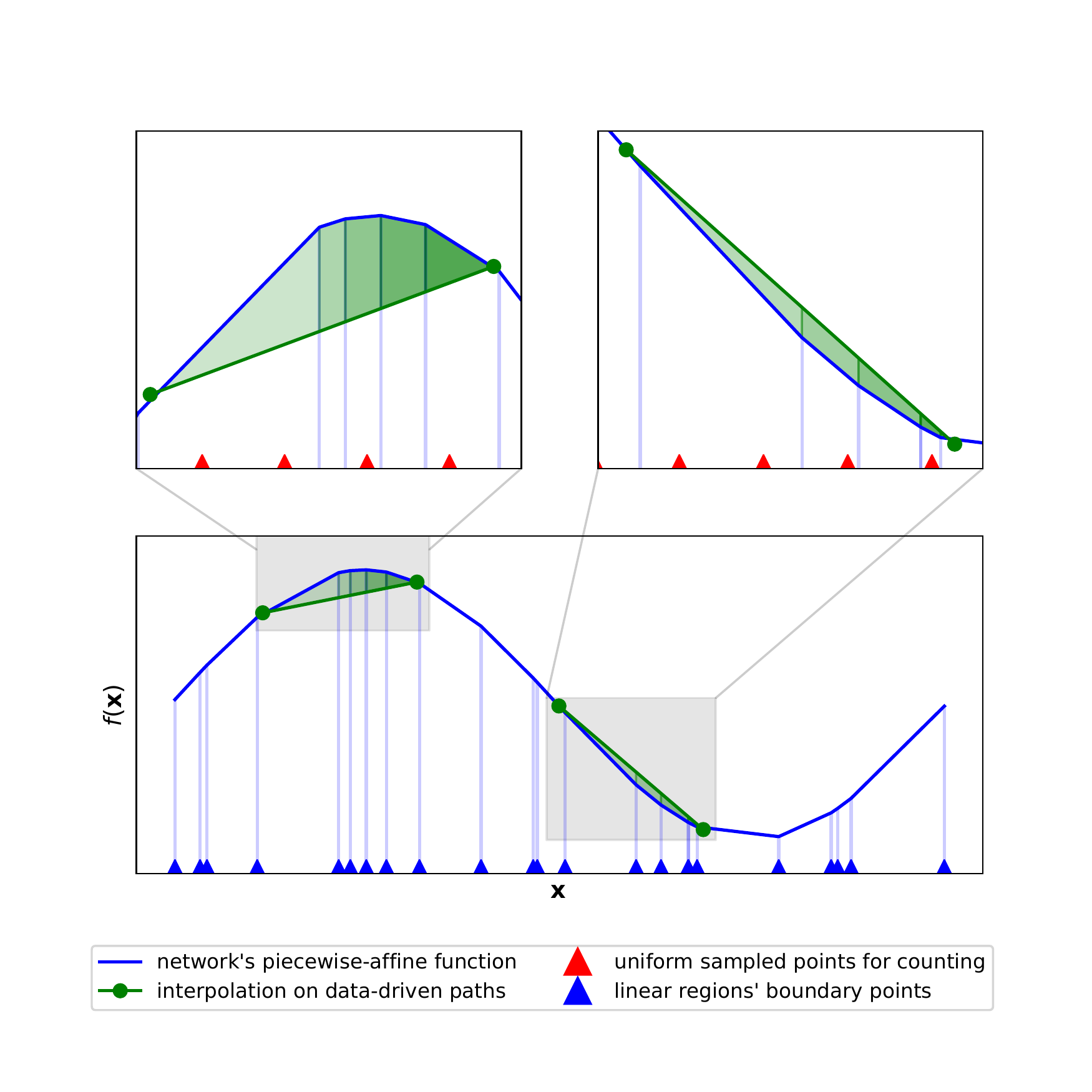}
	\end{center}
	\caption{\small\textbf{General methodology:} 
		we quantify the nonlinearity of continuous piece-wise affine functions (blue line) for ReLU networks, which implicitly partition the input space into disjoint linear regions (bounded by blue triangles). Prior work used density of such regions as a proxy for nonlinearity. We note that 1) the slope and bias of each affine component, as well as the size of each region, affect nonlinearity, and 2) the same number of regions may correspond to different levels of nonlinearity (top box of the figure). Thus, we devise a novel measure based on absolute deviation from affine interpolation (green line) which better captures the non-linearity of learned functions (green shaded area).
		Furthermore, existing numerical methods measure density by sampling equidistant points (red triangles) which can be either prohibitively expensive or miss some regions depending on the granularity of sampling. Here, we propose an adaptive numerical algorithm for accurately counting all regions. Details of the proposed counting algorithm, the novel measure, and corresponding experiments are in sections~\ref{sec:method:discovery}, \ref{sec:method:deviation}, and \ref{sec:experiments} respectively.}
\label{fig:illustration}
\end{figure}
\begin{figure}[t!]
    \begin{center}
        \includegraphics[scale=0.2]{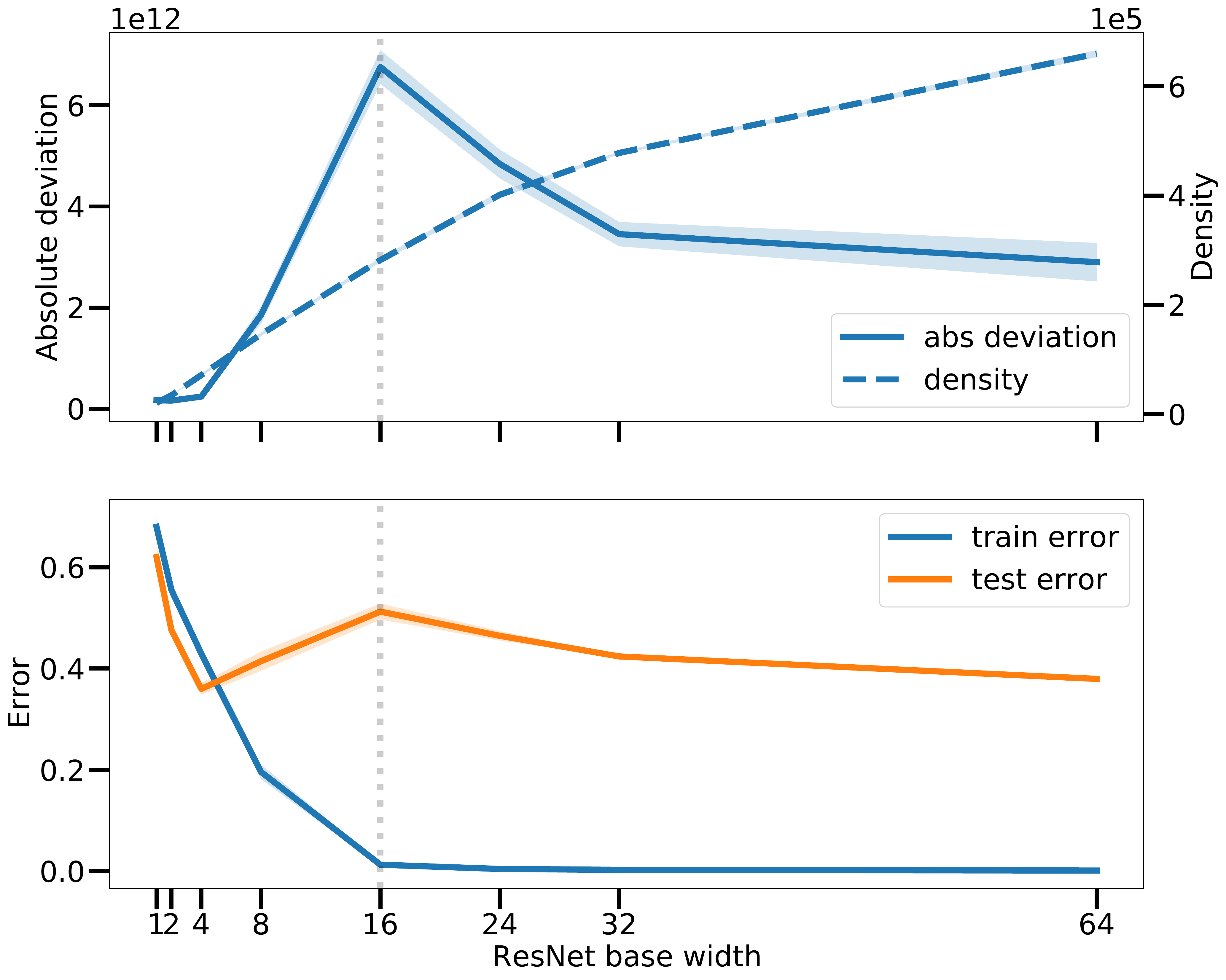}
    \end{center}
    \caption{\textbf{Absolute deviation closely follows the test error past the interpolation threshold in a model-wise double descent regime.} At the same time, linear region density grows monotonically with model size rendering it unsuitable as a proxy for the complexity of a trained ReLU network. Thus, while larger models have in principle higher expressivity, their effective nonlinearity \textit{decreases} during the second descent. (Top) Average median density and absolute deviation (over $3$ seeds) computed on the CIFAR-10 training set with $20\%$ noisy labels, for ResNets18s of increasing base width. (Bottom) Train and test error ($0/1$ loss) as a function of ResNet18 base width. The networks are trained with no explicit regularization or data augmentation, using Adam with base learning rate $1\mathrm{e}-4$.}
    \label{fig:exps:double-descent}
\end{figure}

Estimating the complexity of deep networks trained in practice is an open research problem posing several challenges~\citep{kawaguchi2017generalization, neyshabur2015search}. For a network equipped with piece-wise linear activation functions -- most prominently ReLU -- one avenue for studying complexity is through the lens of \textit{linear regions}, namely the connected components induced on the input space by the piece-wise affine function parameterized by the network~\citep{balestriero2018spline, montufar2014number, pascanu2013number}.

Early works on linear regions highlighted theoretical gains in model expressivity for deep networks, as opposed to wider and shallower ones~\citep{cohen16expressive, telgarsky16benefits, hastad1986optimal}. Later studies mainly focused on estimating the \textit{density} of linear regions, \ie bounding and counting the number of affine components realized by a given network architecture, proposed as a measure for studying the complexity of hierarchical representations~\citep{xiong2020number, hanin2019deep, novak2018sensitivity, serra2018bounding, arora2018understanding, montufar2014number, pascanu2013number}. Intuitively, the main rationale motivating such studies is that, in order for a network to model complex non-linear behaviour, the resulting piece-wise affine function should count many affine components.

Currently, it is debated whether linear region density is the most suited metric for capturing the complexity of piece-wise affine functions parameterized by deep networks~\citep{trimmel2021tropex,lejeune2019implicit, hanin2019complexity}, but there is no systematic study presenting evidence of where density fails. 

Crucially, existing empirical studies of linear regions are limited to small networks, and have thus not explored the relationship between model size -- which directly controls expressivity -- and nonlinearity. In fact, existing numerical methods for density estimation~\citep{novak2018sensitivity} suffer from the limitations of uniform sampling (Figure~\ref{fig:illustration}, top), while exact analytic approaches are unable to scale to large  networks~\citep{zhang2020empirical, hanin2019complexity, hanin2019deep}.

In this work, we systematically investigate whether all linear regions equally contribute nonlinearity relevant to learning for ReLU networks. We expose shortcomings of estimating the complexity of piece-wise affine functions exclusively by the number of their affine components, by contrasting linear region density to a more principled way of estimating nonlinearity. Intriguingly, using our nonlinearity measure, we empirically observe that for increasing model size, nonlinearity increases for overfitting networks and then decreases for large, generalizing ones (Figure~\ref{fig:exps:double-descent}); in line with the recently observed deep double descent phenomenon~\citep{nakkiran2019deep, belkin2019reconciling}.

Our contributions~\footnote{Source code available at \url{https://github.com/magamba/linear-regions}} are summarized as follows.
\begin{itemize}
    \item We propose an adaptive numerical algorithm for discovering linear regions along directions in the input space of trained deep networks (section~\ref{sec:method:discovery}). To our knowledge, ours is the first study estimating accurate linear region density for architectures such as ResNet~\citep{he2015delving}. We present our findings on the CIFAR-10 and CIFAR-100 datasets~\citep{krizhevsky2009learning}.
    \item We contrast linear region density to a more principled way of estimating nonlinearity of piece-wise affine functions expressed by ReLU networks (section~\ref{sec:method:deviation}), accounting for the size of each linear region, as well as nonlinearity expressed by the respective affine components and their bias terms.
    \item Building on previous work~\citep{novak2018sensitivity}, we study directions in the input space that meaningfully capture nonlinearity, but along which linear region density is unsuited for discerning generalizing networks from memorizing ones (section~\ref{sec:experiments}).
    \item We establish an empirical connection among test error, overparameterization, and nonlinearity, for which large models that harmlessly interpolate noise show lower nonlinearity as well as test error than models that harmfully overfit noise (section~\ref{sec:exps:double-descent}).
\end{itemize}

\section{METHODOLOGY}
\label{sec:method}
    We fix notation and describe existing approaches for estimating density of linear regions respectively in sections~\ref{sec:method:notation} and \ref{sec:method:density}. In section~\ref{sec:method:discovery}, we present our linear region discovery algorithm. Finally, we introduce a simple measure of nonlinearity of continuous piece-wise affine functions in section~\ref{sec:method:deviation}.

\subsection{Notation}
\label{sec:method:notation}

We consider ReLU networks as functions $\mathbf{f}: \mathbb{R}^d \to \mathbb{R}^K$, obtained by composing $L$ affine layers, each with parameter matrix $\mathbf{W}^\ell \in \mathbb{R}^{d_{\ell} \times d_{\ell-1}}$ and bias vector $\mathbf{b}^\ell \in \mathbb{R}^{d_\ell}$, for $\ell = 1, \ldots, L$, with the continuous piece-wise affine activation function $\varphi(x) = \max(0, x)$, where $d_0 = d, d_L = K$, and $\varphi$ is applied element-wise. The resulting function, $\mathbf{f}(\mathbf{x}) = \mathbf{W}^L\mathbf{x}^{L-1} + \mathbf{b}^L = \mathbf{W}^L\varphi(\ldots \varphi(\mathbf{W}^1 \mathbf{x} + \mathbf{b}^1)) + \mathbf{b}^L$, is itself continuous piece-wise affine, and implicitly induces a partition $\mathcal{P}$ of the input space $\mathbb{R}^d$ into disjoint convex cells $A_\epsilon \in \mathcal{P}$ -- generally referred to as linear regions -- on which the network computes a single affine function \mbox{$\mathbf{f}(\mathbf{x}) = \mathbf{W}_\epsilon\mathbf{x} + \mathbf{b}_\epsilon$}, $\forall \mathbf{x} \in A_\epsilon$, for $\mathbf{W}_\epsilon \in \mathbb{R}^{K \times d}, \mathbf{b}_\epsilon \in \mathbb{R}^{K}$~\citep{balestriero2018spline, montufar2014number}.

For an input $\mathbf{x}\in A_\epsilon$, the parameters of the corresponding affine component $\mathbf{W}_\epsilon\mathbf{x} + \mathbf{b}_\epsilon$, can be obtained by computing the product of matrices $\mathbf{W}_\epsilon = \prod_{\ell = 1}^L \mathbf{W}_\epsilon^\ell$, as follows. For $\ell = 1, \ldots, L$, each matrix $\mathbf{W}^\ell_\epsilon = \mathbf{W}^\ell \odot \mathbf{M}^\ell_{\mathbf{x}}$ is obtained by multiplying element-wise the weight matrix $\mathbf{W}^\ell$ of layer $\ell$ with the binary mask $\mathbf{M}^\ell_{\mathbf{x}} = \mathbf{M}^\ell(\mathbf{x}^{\ell -1}) = (M_{ij}) \in \{0,1\}^{d_\ell \times d_{\ell -1}}$, representing the activation pattern of ReLU at layer $\ell$ when $\mathbf{x}$ is input to the network, \ie $M_{i,:} = 1 $ if $(\mathbf{W}^\ell\mathbf{x}^{\ell -1} + \mathbf{b}^\ell)_i > 0$, and $M_{i,:} = 0$ otherwise. A similar procedure can be used for the bias parameter $\mathbf{b}_\epsilon = \sum_\epsilon\mathbf{b}_\epsilon^\ell$.

Then, using the indicator function, $\mathbf{f}$ can be decomposed as the sum over all linear regions of its affine components~\citep{rahaman19spectral}, yielding 
\begin{equation}
\label{eq:method:indicator}
\mathbf{f}(\mathbf{x}) = \sum_\epsilon \mathbb{1}_{\mathbf{A}_\epsilon}(\mathbf{x})(\mathbf{W}_\epsilon\mathbf{x} + \mathbf{b}_\epsilon) \quad \forall \mathbf{x} \in \mathbb{R}^d
\end{equation}
and making the locally affine behaviour of $\mathbf{f}$ explicit.

Next, we summarize prior work for estimating linear region density and highlight their limitations.
\subsection{Empirical Estimates of Linear Region Density}
\label{sec:method:density}

\textbf{Computational Complexity.} The density of linear regions, defined as the size of the partition $\mathcal{P}$, has been used as a way to quantify nonlinearity of ReLU networks. In practice, exact calculation of $\mathcal{P}$  involves traversing the network sequentially, neuron by neuron, and computing how the hyperplanes defined by each neuron intersect with those at earlier layers~\citep{zhang2020empirical, hanin2019complexity, novak2018sensitivity}. Hence, estimating global density of overparameterized networks is prohibitively expensive in practice for high-dimensional input domains. Furthermore, the computational cost increases for convolutional layers whose weight tensors need unfolding into sparse weight matrices~\citep{zhang2020empirical}. 

\textbf{Analytical Methods.} To mitigate such computational challenges, prior work have resorted to calculating density of regions on 1-D and 2-D compact subsets $\mathcal{B} \subset \mathbb{R}^d$ of the input space, thus estimating the size of the restriction of $\mathcal{P}$ to $\mathcal{B}$.  Exact analytical methods have so far been bound to small networks~\citep{zhang2020empirical, hanin2019deep, hanin2019complexity} and simple datasets~\citet{hanin2019deep, hanin2019deep}, making numerical approaches appealing. 

\textbf{Numerical Methods.} Numerical methods have focused on bounded 1-dimensional trajectories in the input space, obtained by sampling equally-spaced points $\{\mathbf{x}_0, \ldots, \mathbf{x}_n\}$, with a fixed step size \mbox{$\lambda=\|\mathbf{x}_{i+1} - \mathbf{x}_i \|$}, and exploiting the binary activation pattern $\big{[}\vect(\mathbf{M}^{1}_{\mathbf{x}}), \ldots, \vect(\mathbf{M}^{L}_{\mathbf{x}}) \big{]}$ of each sample $\mathbf{x}$ as a signature for each region $A_{\epsilon_i}$, thus estimating density by the number of unique such patterns~\citep{novak2018sensitivity, raghu2017expressive}.

\textbf{Shortcomings.} Both analytical and numerical strategies attempt to capture nonlinearity of piece-wise affine functions by enumerating regions. On the one hand, the sequential nature of analytic approaches makes them unable to scale to realistic networks. On the other hand, numerical methods based on uniform sampling can only provide estimates on the number of regions by comparing activation patterns, without capturing the geometry of the underlying partition $\mathcal{P}$. Importantly, numerical approaches might potentially miss small regions when the step size $\lambda$ is too large, or become prohibitively expensive for fine-grained sampling. Crucially, all existing strategies based on linear region density intrinsically assume all regions to equally contribute nonlinearity, independently of other properties of the corresponding affine function, as highlighted in Figure~\ref{fig:illustration}.

In the next section, we remove the dependency of numerical counting on a uniform step size $\lambda$, and instead provide a linear region discovery method that captures the size of linear regions along a given direction in the input space. Finally, in section~\ref{sec:method:deviation}, we introduce a simple measure that ties linear regions to nonlinearity of the corresponding piece-wise affine function.

\begin{figure*}[t!]
    \begin{center}
    \resizebox{\textwidth}{!}{%
        \subfloat[]{\includegraphics[width=0.3\linewidth]{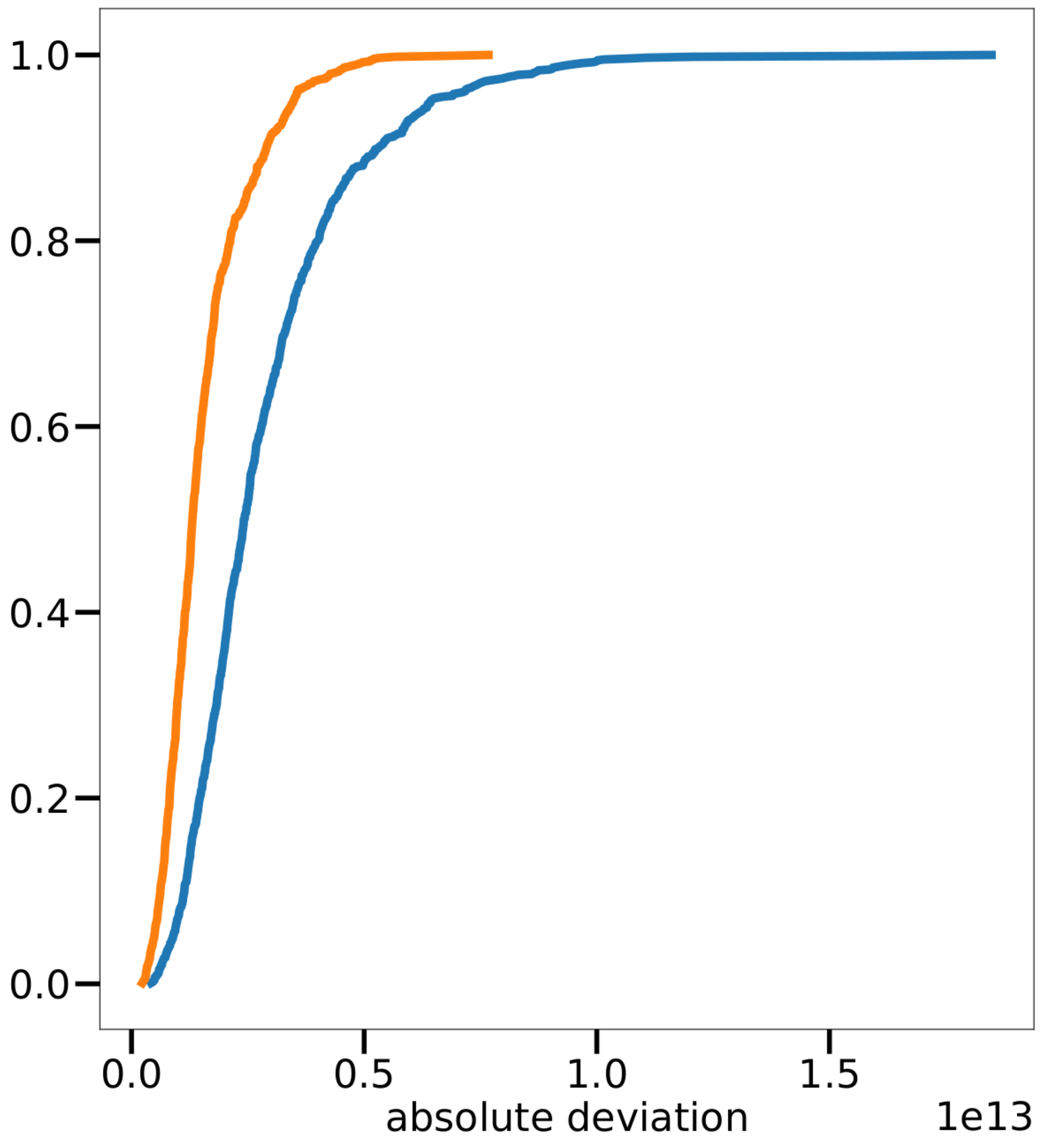}}  
        \subfloat[]{\includegraphics[width=0.3\linewidth]{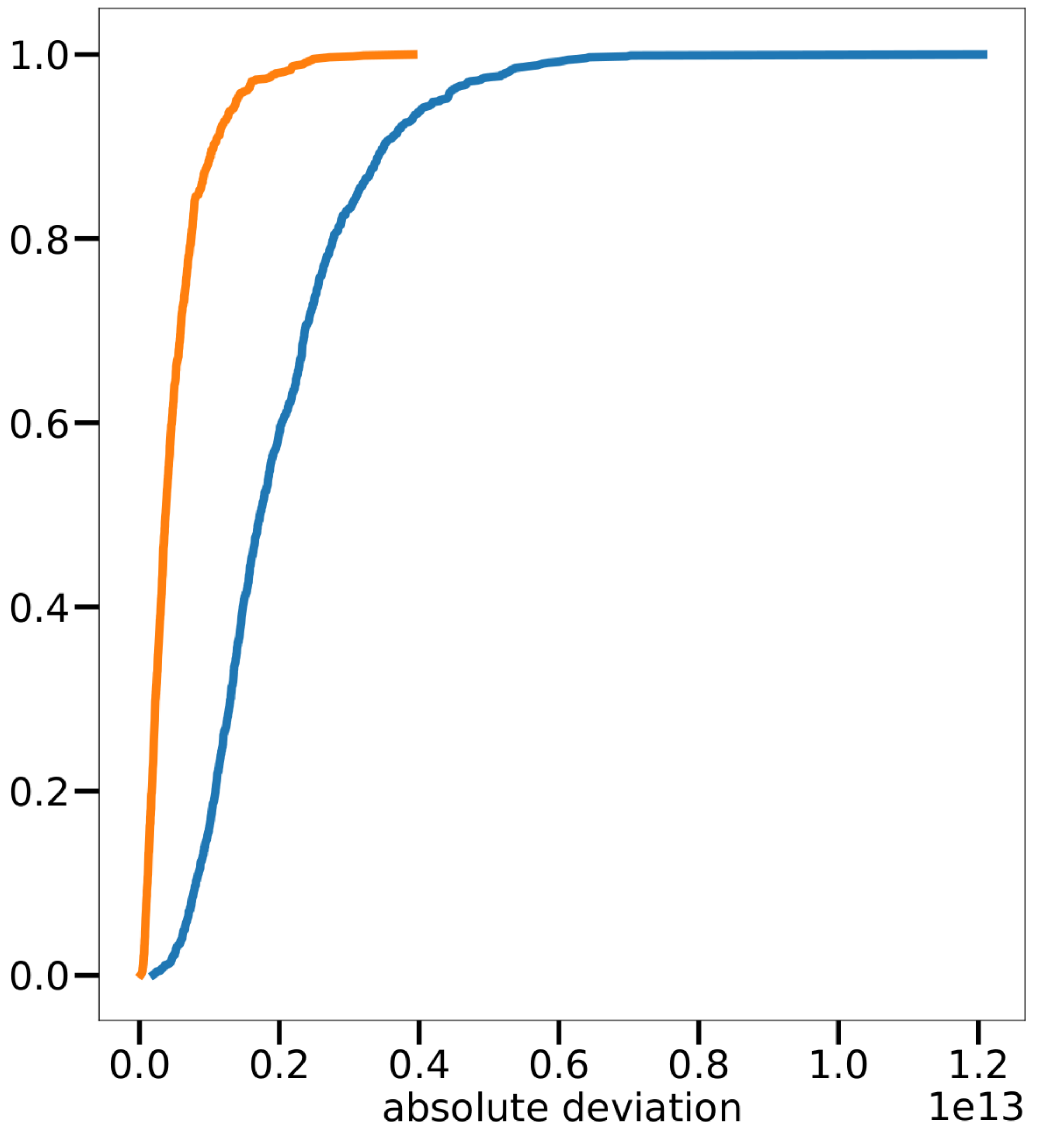}}  
        \subfloat[]{\includegraphics[width=0.295\linewidth]{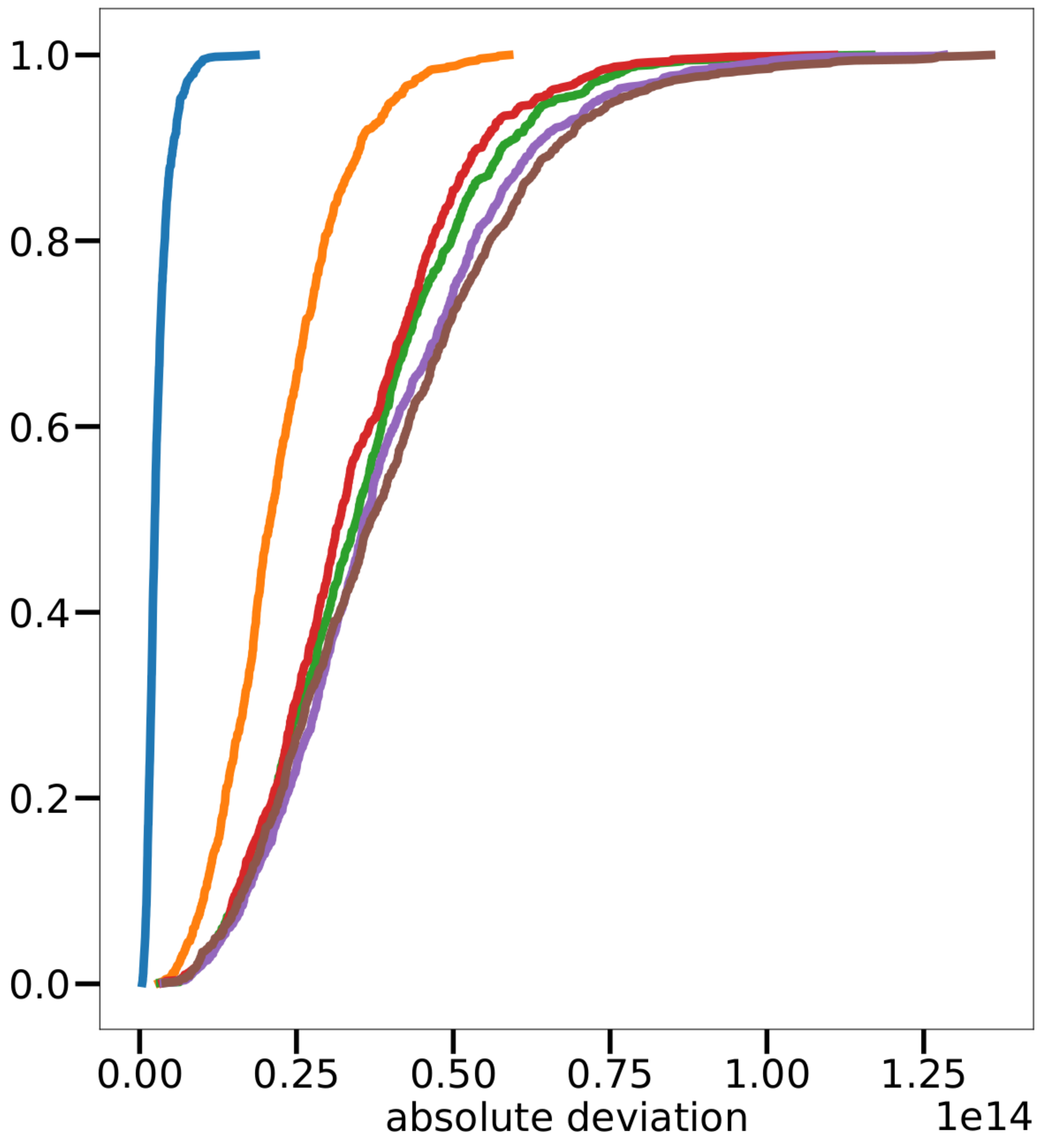}}  
        \subfloat[]{\includegraphics[width=0.3\linewidth]{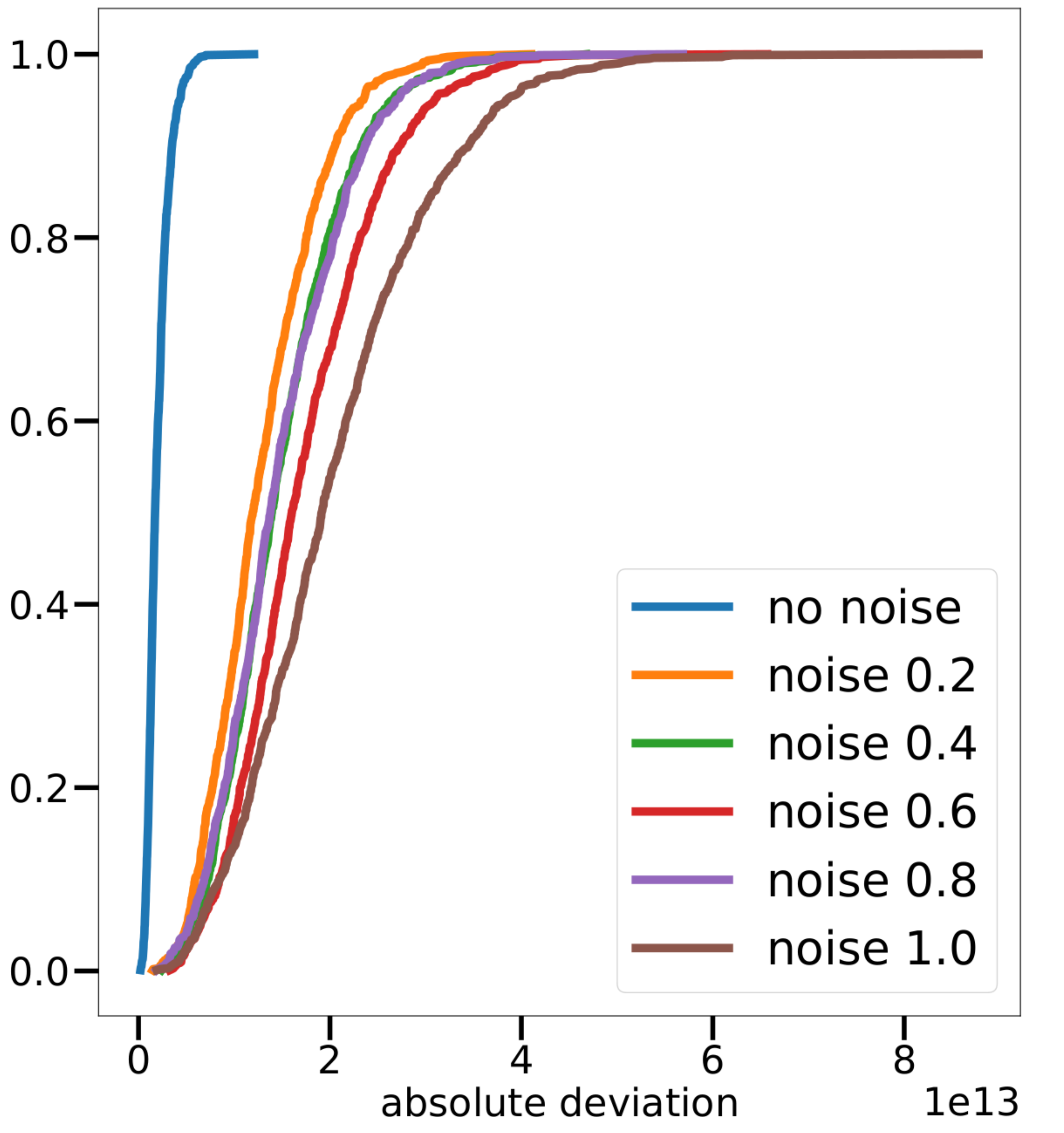}}
    }%
    \end{center}
    \caption{Empirical cumulative distribution functions for absolute deviation along data-driven paths sampled from the CIFAR-10 training set, for several training settings. From the left, a) VGG8, and b) ResNet18, each trained with and without data augmentation. c) VGG8, and d) ResNet18 each trained on CIFAR-10 with increasingly noisy training labels. Absolute deviation captures increased smoothness for networks trained with data augmentation, as well as nonlinearity expressed by networks fitting increasingly harder tasks, thus constituting a principled baseline for studying linear region density in practical settings.}
    \label{fig:exps:dist-population}
\end{figure*}

\subsection{Adaptive Linear Region Discovery}
\label{sec:method:discovery}

In this section, we present an adaptive numerical algorithm for discovering linear regions along a direction $\mathbf{d}$ in the input space, starting from an input point $\mathbf{x}_0$.

We begin by recalling that, for each affine layer $\ell$ with parameters $\mathbf{W}^\ell, \mathbf{b}^\ell$, each neuron $j$ induces a hyperplane $\mathcal{H}_j^\ell : {\mathbf{w}_j^\ell}^T\mathbf{x} + b_j^\ell = 0$ in the preactivation space $\mathbb{R}^{d_{\ell -1}}$ of the layer, with $\mathbf{w}_j^\ell := \mathbf{W}^\ell_{j,:}$ and $b_j^\ell := \mathbf{b}_j^\ell$. In principle, given an input $\mathbf{x}_0$, belonging to linear region $A_\epsilon$, the boundaries of $A_\epsilon$ can be analytically determined as summarized in section~\ref{sec:method:density}. 
Here, we instead propose a faster numerical algorithm finding the closest linear region boundary of $A_\epsilon$ to $\mathbf{x}_0$, along $\mathbf{d}$. 

Let $\mathbf{x}^{\ell-1} = \varphi(\mathbf{W}^{\ell-1}\mathbf{x}^{\ell -2} + \mathbf{b}^{\ell -1})$ denote the output of layer $\ell -1$. Then, for each hyperplane $\mathcal{H}_j^\ell$ induced by layer $\ell$, we solve the linear problem specified by Equation~\ref{eq:method:lambda}, to determine the closest boundary to $\mathbf{x}^{\ell-1}$ at layer $\ell$ in the direction $\mathbf{d}$. Assuming that $\mathbf{x}^{\ell -1}$ lies on the negative side of $\mathcal{H}_j^\ell$~\footnote{If instead $\mathbf{x}^{\ell -1}$ lies on the positive side of $\mathcal{H}_j^\ell$, the inequality is flipped, as well as the sign of $\lambda_j^\ell$.}, the smallest displacement $\lambda^\ell_j$ to cross $\mathcal{H}^\ell_j$ along $\mathbf{d}$ is computed by solving
\begin{align}
\lambda_j^\ell &= \min\limits_{\substack{\lambda \in \mathbb{R} \\ |\lambda| > \tau}} {\mathbf{w}^\ell_j}^T (\mathbf{x}^{\ell -1} + \lambda \mathbf{d}^\ell) + b^\ell_j > 0 \\
               &= \min\limits_{\substack{\lambda \in \mathbb{R} \\ |\lambda| > \tau}} \lambda > \frac{-{\mathbf{w}^\ell_j}^T\mathbf{x}^{\ell -1} -b^\ell_j}{{\mathbf{w}^\ell_j}^T\mathbf{d}^\ell} \label{eq:method:lambda}
\end{align}
where $\mathbf{d}^\ell$ represents the direction $\mathbf{d}$ in the preactivation space $\mathbb{R}^{d_{\ell -1}}$ of layer $\ell$, and is obtained as the product $\prod\limits_{p = 1}^{\ell -1} \mathbf{W}_\epsilon^p \cdot \mathbf{d}$, where the matrices $\mathbf{W}_\epsilon^p$ depend on the activation pattern of linear region $A_\epsilon$. To control numerical stability and guarantee that the nearest linear region boundary is always crossed, the solution is computed for $|\lambda| > \tau$, with $\tau \ll 1$, acting as a sensitivity parameter, ensuring that a minimal step along $\mathbf{d}$ is taken at each iteration of the algorithm.~\footnote{If the size of a linear region along $\mathbf{d}$ is lower than $\tau$, then $\tau$ effectively acts as a step size. In practice, choosing $\tau=1\mathrm{e}-6$ with double precision computation resulted in $2\%$ of linear regions being found with size lower or equal than $\tau$, negligibly affecting our comparisons.}. Moreover, the linear problem has no solution if $\mathbf{d} = \bm{0}$, or $\mathbf{d}$ lies on $\mathcal{H}_j^\ell$, in which case the inequality is discarded.

Given a starting point $\mathbf{x}_0 \in \mathbb{R}^d$, and a direction vector $\mathbf{d}$, the smallest displacement $\lambda = \min\limits_{\substack{1 \le j \le d_{\ell -1}, 1 \le \ell \le L}} \lambda_j^\ell$ is computed by a single forward pass of $\mathbf{x}_0$ and $\mathbf{d}$, solving the linear problem~\ref{eq:method:lambda} iteratively for each layer. 

Pseudocode for our algorithm is included in section~\ref{sec:appendix:algorithm}.

\paragraph{Density of Compact 1-D Domains} Iterating the procedure, starting from $\mathbf{x}_0$, it is possible to find all linear regions along a direction $\mathbf{d}$, which we parameterize as a line path $\bm{\pi} : \mathcal{I} = {[}0, 1{]} \to \mathbb{R}^d$, with $\bm{\pi}(t) = \mathbf{x}_0 + t\mathbf{d}$, for $0 \le t \le 1$. Crucially, by convexity of linear regions, $\mathcal{P}$ induces a partition $\mathcal{P}_\mathcal{I} = \{t_\epsilon \in \mathcal{I} : 0 = t_0 < \ldots < t_D = 1\}$ of $\mathcal{I}$, dividing $\mathcal{I}$ into intervals $\mathcal{J}_\epsilon = {[}t_\epsilon, t_{\epsilon + 1} {]}$, with $|\mathcal{J}_\epsilon| = \lambda_\epsilon$ corresponding to the length of region $A_\epsilon$ along $\mathbf{d}$.

Importantly, this method allows us to compute the entry and exit point of $\bm{\pi}$ into each linear region along $\mathbf{d}$, and returns exactly all linear regions of size greater than $\tau$. In the following, we use such quantities to introduce a simple measure of nonlinearity of a network.

\subsection{Quantifying Nonlinearity}
\label{sec:method:deviation}

In principle, as illustrated in Figure~\ref{fig:illustration}, nonlinearity of continuous piece-wise affine functions is controlled by the slope and bias of each component, as well as the volume of the corresponding linear region. Crucially, for overparameterized networks, several neighbouring regions could encode approximately the same affine function, making density unsuited for capturing non-linearity, interpreted as effective complexity.

To investigate the effectiveness of density at capturing nonlinearity of ReLU networks, we hereby introduce a simple notion of nonlinearity along a direction -- called absolute deviation -- measuring the deviation of a ReLU network $\mathbf{f}$ from a simple function $\mathbf{a}$ interpolating the affine components respectively applied by $\mathbf{f}$ at the endpoints of a line path $\bm{\pi}$ in the input space.

For two distinct inputs $\mathbf{x}_0, \mathbf{x}_1 \in \mathbb{R}^d$, respectively falling into linear regions $A_0$ and $A_1$, with corresponding affine transformations $\mathbf{f}_0(\mathbf{x})=\mathbf{W}_0\mathbf{x} + \mathbf{b}_0$, and $\mathbf{f}_1(\mathbf{x})=\mathbf{W}_1\mathbf{x} + \mathbf{b}_1$, we define the affine function $\mathbf{a}(\mathbf{x})$ obtained by linearly interpolating the functions $\mathbf{f}_0$ and $\mathbf{f}_1$, so that, along a line path $\bm{\pi}$ parameterizing $\mathbf{d} = \mathbf{x}_1 - \mathbf{x}_0$, it holds $\mathbf{a}(\mathbf{x}) = \mathbf{f}(\mathbf{x}_0) + t (\mathbf{f}(\mathbf{x}_1) - \mathbf{f}(\mathbf{x}_0))$. That is, we only specify $\mathbf{a}$ so that it interpolates $\mathbf{f}(\mathbf{x}_0)$ and $\mathbf{f}(\mathbf{x}_1)$ along $\mathbf{d}$. Importantly, the difference $\mathbf{f}(\mathbf{x}_1) - \mathbf{f}(\mathbf{x}_0)$ is in general a nonlinear affine function, \ie with non-zero bias term.

Then, for each output dimension $f^k$ of the network, for $k = 1, \ldots, K$, we can measure the nonlinearity of $f^k$ along a line path $\bm{\pi}$ by computing
\begin{equation}
\label{eq:method:deviation}
\begin{aligned}
    & \int_{\bm{\pi}} | f^k(\mathbf{x}) - a^k(\mathbf{x})| d\mathbf{x} \\
  = & \sum\limits_{\epsilon = 0}^{D -1} \int_{t_\epsilon}^{t_{\epsilon +1}} |f^k(\bm{\pi}(t)) - a^k(\bm{\pi}(t)) | \cdot \|\dot{\bm{\pi}}(t) \| dt \\
  = & \sum\limits_{\epsilon = 0}^{D -1} \int_{t_\epsilon}^{t_{\epsilon +1}} |f^k(\mathbf{x}_0 + t\mathbf{d}) - a^k(\mathbf{x}_0 + t\mathbf{d}) | \cdot \|\mathbf{d} \| dt \\
  = & \|\mathbf{d} \| \cdot \sum\limits_{\epsilon = 0}^{D -1} \int_{t_\epsilon}^{t_{\epsilon +1}} \big(|f_\epsilon^k(\mathbf{x}_0) - f_0^k(\mathbf{x}_0) ~+ \\
  ~ & \quad t \big( f_\epsilon^k(\mathbf{x}_1) - f_\epsilon^k(\mathbf{x}_0) - f_1^k(\mathbf{x}_1) + f_0^k(\mathbf{x}_0) \big)|\big) dt 
\end{aligned}
\end{equation}
where $f_\epsilon^k(\mathbf{x}_i) := (\mathbf{W}_\epsilon)_{k, :}~\mathbf{x}_i + (\mathbf{b}_\epsilon)_k$, \ie the $k$-th output logit of the affine component $\epsilon$ of $\mathbf{f}$, evaluated at $\mathbf{x}_i$. A full derivation is presented in section~\ref{sec:appendix:abs_deviation}.

Referring again to Figure~\ref{fig:illustration}, we observe that, 1) since $t_{\epsilon +1} - t_\epsilon = \lambda_\epsilon$, absolute deviation (shaded green area) precisely follows the piece-wise affine structure of $\mathbf{f}$, taking into account the length of linear regions $A_\epsilon$ along $\mathbf{d}$, as well as the slope of $\mathbf{f}$; 2) absolute deviation takes into account all linear regions along a trajectory $\bm{\pi}$, providing an accurate benchmark of expressivity against linear region density; 3) in contrast to total variation measures~\citep{novak2018sensitivity}, the integrand depends on the bias of $\mathbf{f}$, providing a more precise approach than solely estimating any $\mathbf{W}_\epsilon$ through gradient information $\nabla_{\mathbf{x}}\mathbf{f}$, as nonlinearity arising from bias terms is otherwise lost; 4) the integrand smoothly interpolates between the affine function $f^k_0$ and $f^k_1$ such that, if $\mathbf{f}$ computes affine functions that are approximately similar to $\mathbf{f}(\mathbf{x}_0)$ or $\mathbf{f}(\mathbf{x}_1)$ along $\bm{\pi}$, then the resulting absolute deviation will be low; 5) if $\mathbf{x}_0, \mathbf{x}_1$ are training or validation points, then $\bm{\pi}$ anchors $\mathbf{a}$ to the support of the data-generating distribution at $\mathbf{x}_0$ and $\mathbf{x}_1$, and allows for computing nonlinearity in proximity of the where the data lies in the input space.

Lastly, we note that absolute deviation is a one-dimensional estimate of nonlinearity, meant to contrast density of linear regions, rather than to stand as a novel generalization measure. In particular, its effectiveness at capturing nonlinearity expressed by learning is bound to evaluating the measure along meaningful directions $\mathbf{d}$ in the input space. In section~\ref{sec:experiments}, we exploit data-driven trajectories introduced in~\citep{novak2018sensitivity}, which were shown to capture sensitivity of $\mathbf{f}$ to input perturbations.

\section{EXPERIMENTS}
\label{sec:experiments}
    In this section, we empirically investigate practical training settings in which density of linear regions is an unreliable estimator of nonlinearity. We apply our linear region discovery algorithm to $N = 1024$ closed paths $\bm{\pi}_n$ in the input space of trained networks, constructed by connecting a training sample $\mathbf{x}_n^0$, with augmented versions obtained by deterministically translating $\mathbf{x}_n^0$ along a circular trajectory~\citep{novak2018sensitivity}, which we define of radius $4$ to reflect common data-augmentation strategies. Details on the construction are presented in section~\ref{sec:appendix:paths}.  Each closed path is defined using $A = 8$ augmented samples, each connected to the next one along the circular trajectory, by straight lines $\mathbf{d}_n^a := \mathbf{x}_n^{(a+1) \%A} - \mathbf{x}_n^a$, for $a = 0, \ldots A$. Each point $\mathbf{x}_n^a$ anchors the path to the support of the data distribution, ensuring that density and deviation along each line $\mathbf{d}_n^a$ are evaluated in proximity of the data manifold in pixel space. Starting from a base point $\mathbf{x}_n^0$ for each path $\bm{\pi}_n$, we compute the entry and exit point of $\bm{\pi}_n$ for each linear region it crosses, and compute linear region density as well as absolute deviation, for several trained networks.

We perform our experiments on the CIFAR-10 and CIFAR-100 datasets~\citep{krizhevsky2009learning}, using a VGG-like network~\citep{simonyan15very} with 8 layers, and a ResNet18~\citep{he2015delving} with base width $16$. All measures are computed on the training split of the dataset considered. All networks fitting noisy labels are trained until $100\%$ training accuracy is reached. Our experimental setup and network architectures are detailed in sections~\ref{sec:appendix:setup} and~\ref{sec:appendix:models}. 

Finally, we recall that, for a network with $K$-dimensional output, absolute deviation produces $K$ scalar scores -- one per logit. Throughout our experiments, we interpret such scores together as a $K$-dimensional vector, of which we take the $\ell_2$ norm to produce a single scalar measure per path, which can be contrasted to linear region density along the same path. We keep the set of paths fixed throughout our experiments, in order to study how our measures vary when the training setup is changed.

\subsection{Absolute Deviation Captures Complexity of Trained Networks}
\label{sec:exps:deviation}

\begin{figure}[t!]
    \begin{center}
        \subfloat[\label{fig:exp:cdf-compare-vgg}]{\includegraphics[width=0.5\linewidth]{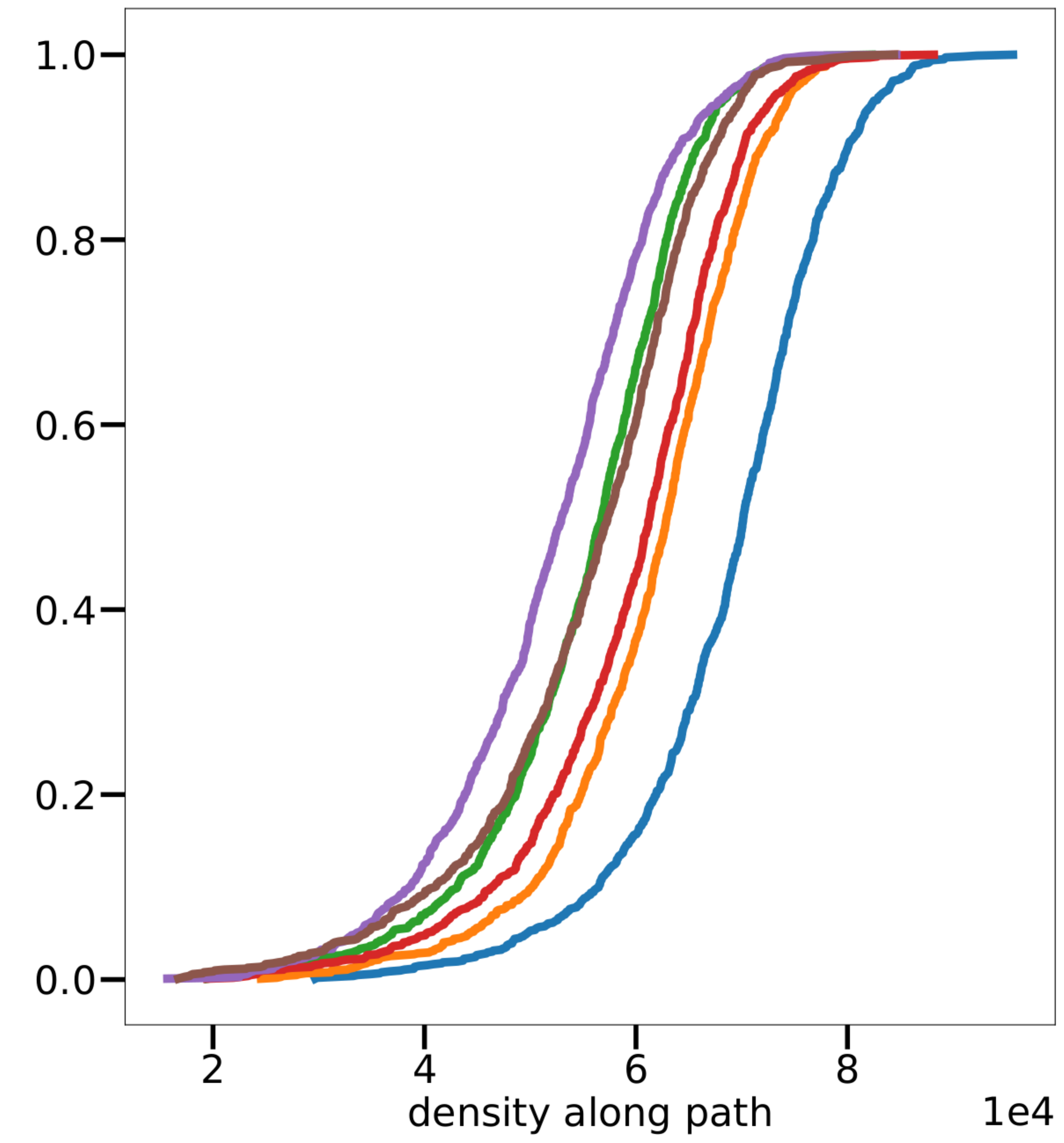}} ~
        \subfloat[\label{fig:exp:cdf-compare-resnet}]{\includegraphics[width=0.49\linewidth]{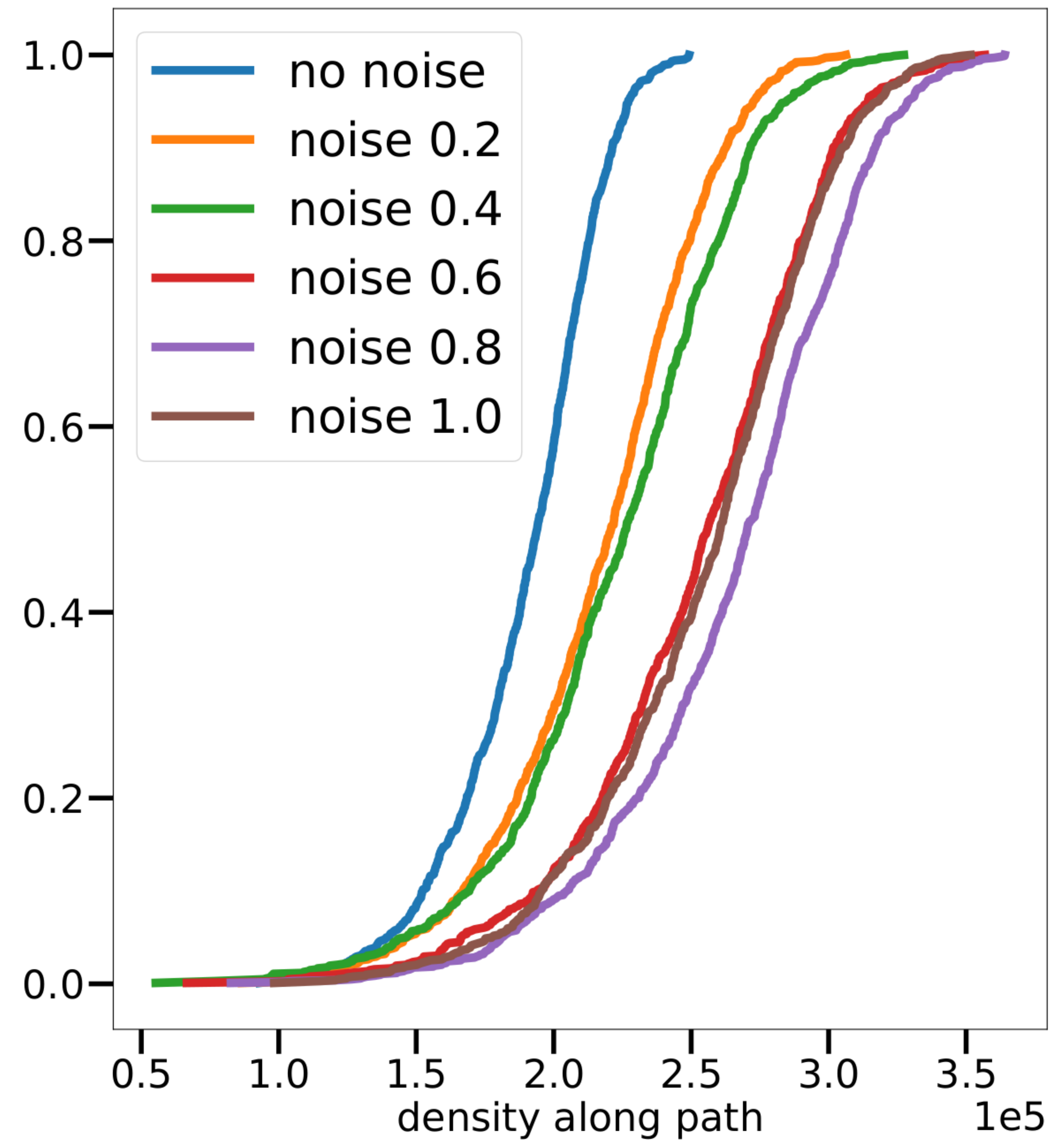}}
    \end{center}
    \caption{ECDF of density along data-driven paths on CIFAR-10 samples, for increasingly noisy labels. Importantly, it can be observed that a fixed density value can realize functions of different nonlinearity. a) VGG8. b) ResNet18.}
    \label{fig:exps:task-ranking}
\end{figure}

We begin by evaluating whether absolute deviation is a suitable measure of nonlinearity, by studying networks trained on learning tasks with increasingly noisy training labels, pushing each network towards learning increasingly nonlinear decision functions. In Figure~\ref{fig:exps:dist-population}, we compare the functions parameterized by VGG8 and ResNet18 on CIFAR-10 for each training setting, by studying the Empirical Cumulative Distribution Function (ECDF) of absolute deviation evaluated on the $N$ paths. Furthermore, we evaluate whether absolute deviation is able of capturing smoothness of networks in the input variable $\mathbf{x}$, as promoted by data augmentation. For both VGG and ResNet, in the noisy labels setting, we observe that absolute deviation is concentrated towards relatively low values for networks trained on clean labels, while higher deviation is realized by networks fitting harder tasks, with the $100\%$ noisy labels setting realizing highest deviation. Furthermore absolute deviation clearly separates piece-wise affine functions regularized with data augmentation from those trained in vanilla settings with no augmentation. Therefore, we conclude that our novel metric is a suitable candidate for assessing the ability of linear region density to capture nonlinearity for networks trained in practice.

For comparison, Figure~\ref{fig:exps:task-ranking} shows the ECDF for density along the same paths and training settings. Absolute deviation better separates noisy labels training settings for clean labels, by concentrating probability mass towards low values for less-noisy datasets. Furthermore, while VGG8 realizes higher density for clean labels than for any noisy setting, ResNet18 expresses more regions for noisy labels, showing that density is an unreliable predictor of expressivity.

\subsection{Absolute Deviation Better Ranks Nonlinearity}
\label{sec:exps:complexity}

\begin{figure}[t]
    \begin{center}
        \includegraphics[width=\linewidth]{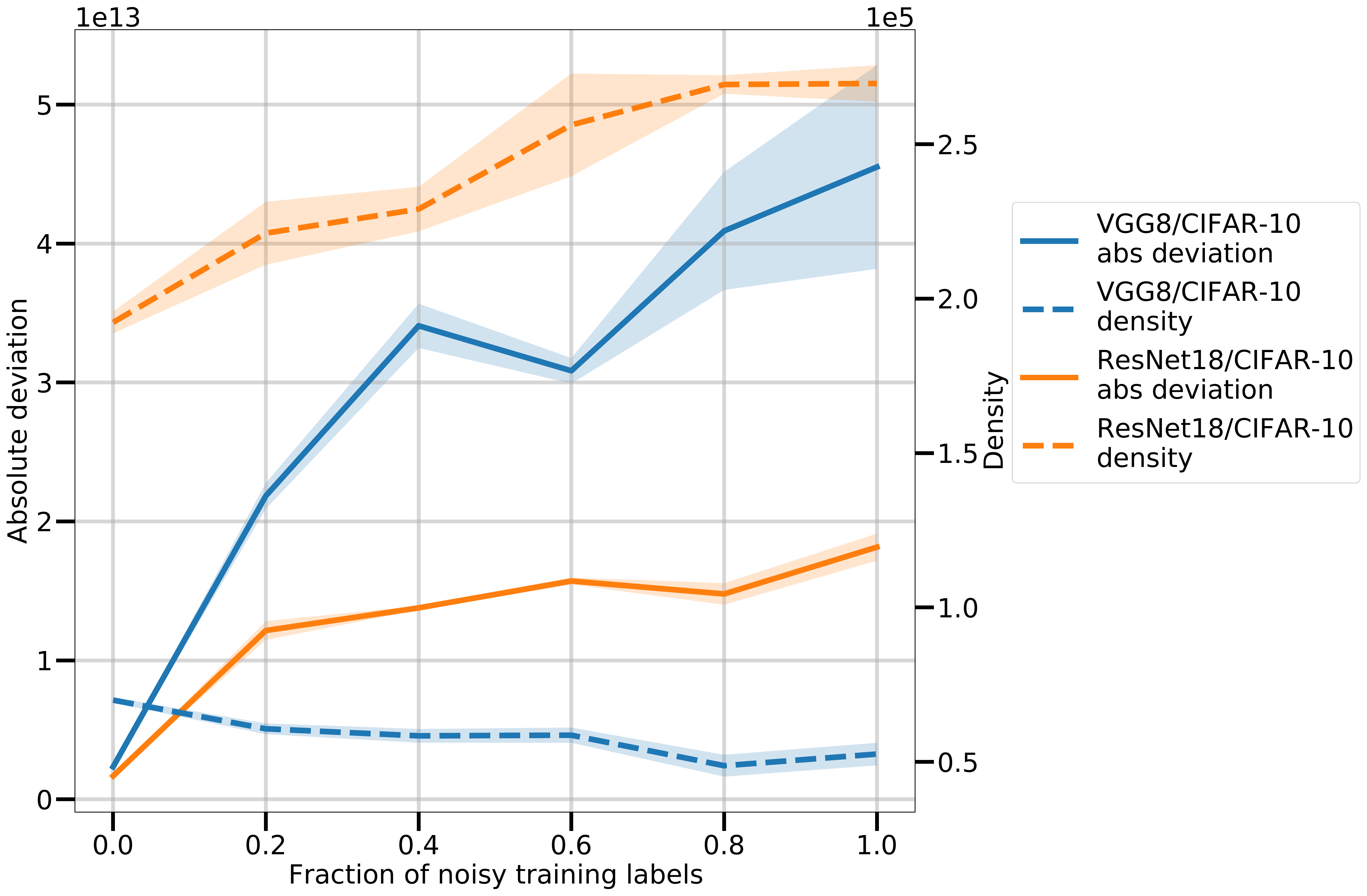}
    \end{center}
    \caption{Median density and absolute deviation for increasingly noisy learning tasks, averaged over three independent runs. While absolute deviation almost always increases for harder problems, density decreases for ResNet18 and increases for VGG8, thus providing a noisy predictor of nonlinearity.}
    \label{fig:exps:task-complexity}
\end{figure}

Next, we investigate how absolute deviation and density fare at ranking learning tasks of increasing complexity, represented by increasing fraction of noisy training labels on CIFAR-10. Figure~\ref{fig:exps:task-complexity} shows the average median density and median absolute deviation, as a function of the fraction of noisy labels, with each median averaged over three independent training runs. Absolute deviation correctly separates networks trained on clean labels from all noisy labels setting, and correctly ranks networks trained on $80\%$ and $100\%$ noisy labels as the most nonlinear, while at the same time fails to separate between $40\%$ and $60\%$ noise for VGG, and $60\%$ and $80\%$ for ResNet. Strikingly, density increases with task complexity for ResNet18, but decreases for VGG8, showing that, along the data-driven directions considered, density is a fragile estimator of nonlinearity, and that a more precise measure of variation of the piece-wise affine function itself offers a more robust measure.

\subsection{Density Poorly Correlates with Absolute Deviation}
\begin{table}[]
\caption{Spearman rank correlation between density and absolute deviation along data-driven paths. Consistently, density poorly correlates with absolute deviation, sometimes anticorrelating.}
\begin{center}
\resizebox{0.9\linewidth}{!}{%
\begin{tabular}{@{}llll@{}}
\toprule
             & \multicolumn{2}{c}{CIFAR-10}                             & \multicolumn{1}{c}{CIFAR-100} \\ \midrule
             & \multicolumn{1}{c}{VGG8} & \multicolumn{1}{c}{ResNet18} & \multicolumn{1}{c}{VGG8}     \\
vanilla      & -0.15 $\pm$ 0.00          & 0.05 $\pm$ 0.03              & 0.00 $\pm$ 0.02               \\
augment.     & -0.08 $\pm$ 0.02          & 0.17 $\pm$ 0.06              & 0.11 $\pm$ 0.08               \\
noise 0.2    & 0.04 $\pm$ 0.02           & 0.21 $\pm$ 0.02              &                               \\
noise 0.4    & 0.11 $\pm$ 0.02           & 0.21 $\pm$ 0.04              &                               \\
noise 0.6    & 0.18 $\pm$ 0.07           & 0.25 $\pm$ 0.03              &                               \\
noise 0.8    & 0.26 $\pm$ 0.04           & 0.22 $\pm$ 0.02              &                               \\
noise 1.0    & 0.14 $\pm$ 0.04           & 0.27 $\pm$ 0.04              &                               \\ \bottomrule
\end{tabular}
}%
\end{center}
\label{tab:exps:spearman}
\end{table}

Recalling that absolute deviation is expressed as a sum of non-negative terms over linear regions (Equation~\ref{eq:method:deviation}), we investigate whether it can simply be explained by density. For ResNet18 and VGG8 trained on CIFAR-10 and CIFAR-100, we compute the Spearman rank correlation coefficient between density and deviation, across $N$ fixed data-driven paths. Table~\ref{tab:exps:spearman} reports the average correlation coefficients with one standard deviation, computed over three independent training runs for each network and training setting. Consistently, density correlates poorly with absolute deviation, presenting negative correlations for VGG trained without data augmentation on both datasets.

\subsection{Density May Fail to Detect Increased Nonlinearity}
\begin{table*}[]
\caption{Paired differences for several training settings, averaged over three independent training runs for each training setting.}
\begin{center}
\resizebox{0.9\textwidth}{!}{%
\begin{tabular}{@{}lllllll@{}}
\toprule
                          & \multicolumn{4}{c}{CIFAR-10}                                          & \multicolumn{2}{c}{CIFAR-100}     \\ \midrule
                          & \multicolumn{2}{c}{VGG8}         & \multicolumn{2}{c}{ResNet18}      & \multicolumn{2}{c}{VGG8}         \\
\multicolumn{1}{c}{} &
  \multicolumn{1}{c}{Abs deviation} &
  \multicolumn{1}{c}{Density} &
  \multicolumn{1}{c}{Abs deviation} &
  \multicolumn{1}{c}{Density} &
  \multicolumn{1}{c}{Abs deviation} &
  \multicolumn{1}{c}{Density} \\
vanilla vs augment. &
  0.98 $\pm$ 0.01 &
  0.66 $\pm$ 0.40 &
  1.00 $\pm$ 0.00 &
  0.03 $\pm$ 0.04 &
  1.00 $\pm$ 0.00 &
  0.38 $\pm$ 0.24 \\
noise 0.2 vs no noise       & 1.00 $\pm$ 0.00 & 0.00 $\pm$ 0.00 & 1.00 $\pm$ 0.00 & 0.95 $\pm$ 0.02 &                 &                 \\
noise 0.4 vs no noise       & 1.00 $\pm$ 0.00 & 0.00 $\pm$ 0.00 & 1.00 $\pm$ 0.00 & 0.78 $\pm$ 0.02 &                 &                 \\
noise 0.6 vs no noise       & 1.00 $\pm$ 0.00 & 0.00 $\pm$ 0.01 & 1.00 $\pm$ 0.00 & 0.86 $\pm$ 0.03 &                 &                 \\
noise 0.8 vs no noise       & 1.00 $\pm$ 0.00 & 0.00 $\pm$ 0.00 & 0.99 $\pm$ 0.00 & 0.91 $\pm$ 0.02 &                 &                 \\
noise 1.0 vs no noise       & 1.00 $\pm$ 0.00 & 0.00 $\pm$ 0.00 & 1.00 $\pm$ 0.00 & 1.00 $\pm$ 0.00 &                 &                 \\ 
vanilla trained vs init.    & 1.00 $\pm$ 0.00 & 1.00 $\pm$ 0.00 & 1.00 $\pm$ 0.00 & 0.79 $\pm$ 0.05 & 1.00 $\pm$ 0.00 & 1.00 $\pm$ 0.00 \\
augment. trained vs init.   & 1.00 $\pm$ 0.00 & 1.00 $\pm$ 0.00 & 1.00 $\pm$ 0.00 & 0.99 $\pm$ 0.01 & 1.00 $\pm$ 0.00 & 1.00 $\pm$ 0.00 \\
noise 0.2 trained vs init.  & 1.00 $\pm$ 0.00 & 1.00 $\pm$ 0.00 & 1.00 $\pm$ 0.00 & 0.82 $\pm$ 0.04 &                 &                 \\
noise 0.4 trained vs init.  & 1.00 $\pm$ 0.00 & 1.00 $\pm$ 0.00 & 1.00 $\pm$ 0.00 & 0.96 $\pm$ 0.01 &                 &                 \\
noise 0.6 trained vs init.  & 1.00 $\pm$ 0.00 & 0.99 $\pm$ 0.01 & 1.00 $\pm$ 0.00 & 0.99 $\pm$ 0.01 &                 &                 \\
noise 0.8 trained vs init.  & 1.00 $\pm$ 0.00 & 0.96 $\pm$ 0.03 & 1.00 $\pm$ 0.00 & 1.00 $\pm$ 0.00 &                 &                 \\
noise 1.0 trained vs init.  & 1.00 $\pm$ 0.00 & 0.96 $\pm$ 0.01 & 1.00 $\pm$ 0.00 & 0.94 $\pm$ 0.01 &                 &                
\end{tabular}
}%
\label{tab:exps:differences}
\end{center}
\end{table*}

We now move from a distribution-level study to an instance-based analysis, investigating how density and absolute deviation change at the level of individual data-driven paths, when comparing pairs of training settings. Specifically, for a set of $N$ paths $\{\bm{\pi}_n\}_{n=1}^N$, and a pair of training settings $(T_1, T_2)$, we collect corresponding measures of deviation $\{\dev^i_n\}_{n=1}^N$ and density $\{\den^i_n \}_{n=1}^N$, for $i = 1, 2$, and compute paired differences $\dev^2_n - \dev^1_n$, $\den^2_n - \den^1_n$, for each $n = 1, \ldots, N$. The, for pairs of training settings $(T_1, T_2)$ sorted so that $T_2$ entails learning a function with higher nonlinearity than $T_1$, the ability of density and absolute deviation of capturing nonlinearity at the level of individual paths can be measured by the fraction of positive paired differences observed, \ie $|\{ \dev^2_n - \dev^1_n > 0 ~\text{for}~n = 1, \ldots, N \}|$. Table~\ref{tab:exps:differences} collects our experimental findings, averaged over three independent training runs for each setting.

First, we observe that deviation is able to reliably and consistently detect increased nonlinearity at the level of individual paths, capturing additional nonlinearity expressed by networks when going from no noisy labels to any amount of noisy labels, as well as when disabling data augmentation. In contrast, density fails to do the same, in the same settings, on both CIFAR-10 and CIFAR-100. Second, when comparing each trained network to the same network at initialization, across all training settings, both deviation and density are able to capture nonlinearity arising from learning, showing that the density of linear regions is on average higher for trained networks. However, even in this setting, density fails to capture nonlinearity for up to $20\%$ paths, proving to be a fragile measure in practice. 

\subsection{Density Fails to Distinguish Piece-Wise Affine Functions}
\label{sec:exps:deviation}

\begin{figure}[t!]
    \begin{center}
        \includegraphics[width=0.95\linewidth]{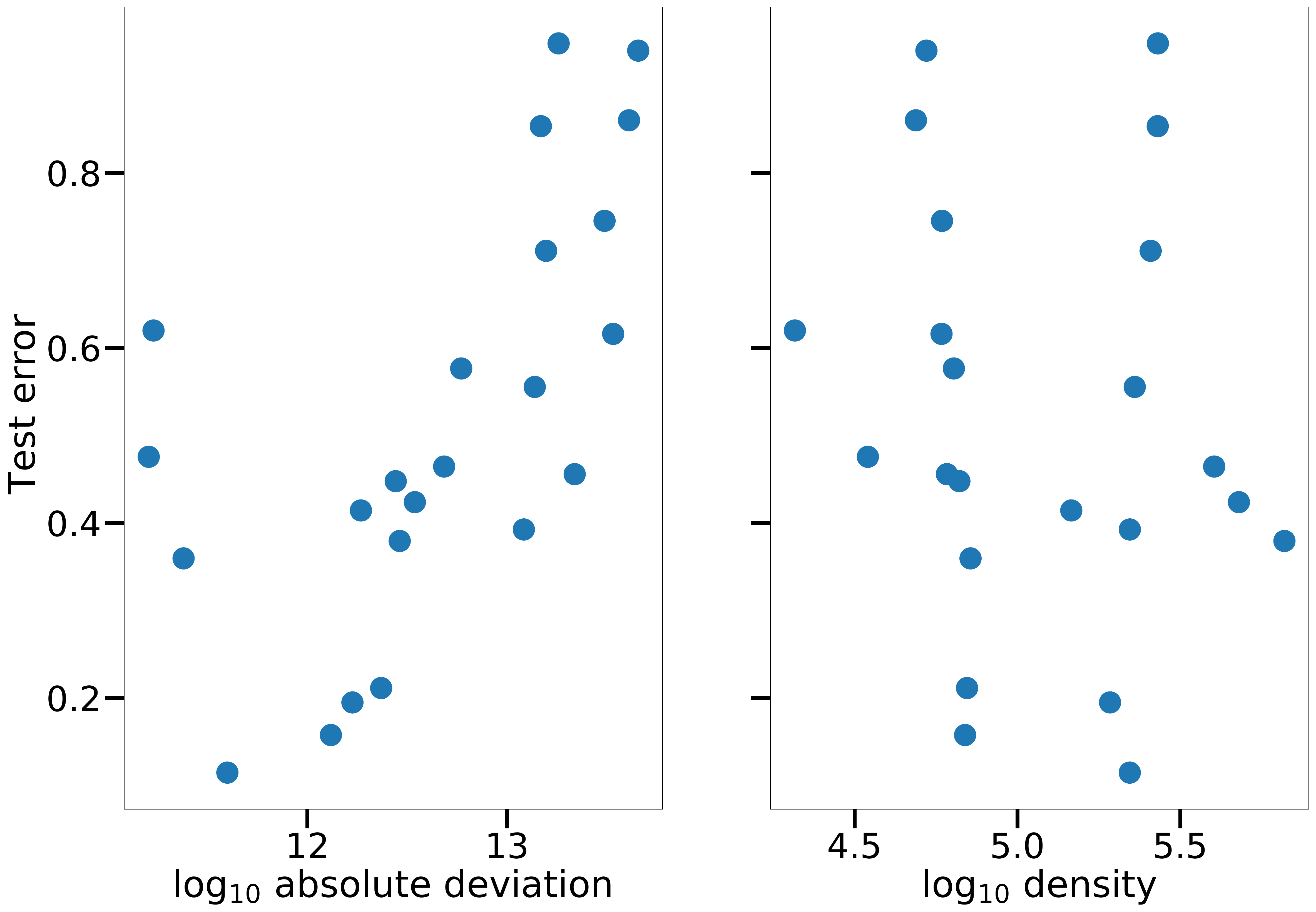}
    \end{center}
    \caption{Mean test error against mean absolute deviation (left) and mean density (right), computed on the training set, for all our experimental settings. Means are computed over three independent runs ($x$-axis shown in log-scale).}
    \label{fig:exps:test_error}
\end{figure}

Next, we investigate whether density and absolute deviation, measured on training data, can predict the network's test error. In Figure~\ref{fig:exps:test_error}, we begin by collecting the test error for all networks, datasets, and training configurations considered in this study. In line with what reported by \citet{novak2018sensitivity}, extending their result to convolutional and residual networks, we observe that measures of variation, like absolute deviation, can better predict test error, here measured using the $0/1$ loss. Importantly, we observe how any fixed value of density can result in networks of vastly different test performance, showing how merely counting the number of affine components is unable to measure effective complexity of continuous piece-wise affine functions parameterized by ReLU networks.

\subsection{Overparameterization Reduces Effective Nonlinearity}
\label{sec:exps:double-descent}

To further explore the connection between nonlinearity and test error, we note that on the one hand, higher expressivity is in principle afforded by increased model size~\citep{telgarsky16benefits,montufar2014number}, while on the other hand overparameterization can promote regularization, observed in the form of reduced test error, in model-wise deep double descent regimes~\citep{nakkiran2019deep, belkin2019reconciling}.

We reproduce the experimental setting of \citet{nakkiran2019deep} (\cfr Figure~4b), and study ResNet18s of increasing base widths $w$ up to $w=64$ (standard ResNet18), on CIFAR-10 with $20\%$ noisy training labels. All networks are trained for $400$ epochs, using Adam with base learning rate $1\mathrm{e}-4$, without any explicit regularization or any data augmentation. We report absolute deviation and density in Figure~\ref{fig:exps:double-descent} (top) as a function of model size, as well as the train and test error (bottom), measured with the $0/1$ loss.

In line with \citet{nakkiran2019deep}, ResNets18s of increasing width first realize decreasing train and test error ($w \le 4$), then overfit the noisy training set incurring in increased test error ($4 < w \le 16$), up to an \textit{interpolation threshold} ($w = 16$), at which zero train error is achieved, with highest test error. Then, a second descent of the test error occurs ($w > 16$), while the networks still perfectly interpolate the training data. Intriguingly, absolute deviation is low for underfitting networks ($w \le 4$, and train error $> 30\%$), which fail to interpolate noise; then the measure increases, peaking at the interpolation threshold, to finally decrease again following the second descent of the test error.

In stark contrast, density monotonically increases with model size. Taken together with absolute deviation, this finding shows that, while larger networks can afford greatly many linear regions, overparameterization past the interpolating threshold induces regularization in the form of reduced nonlinearity, \ie increased local similarity of linear regions. To our knowledge, this is the first experiment explicitly connecting reduced test error in the model-wise double descent regime with reduced nonlinearity of the learned function.

\section{RELATED WORK}
\label{sec:rel_work}
    \paragraph{Density of Linear Regions}

Linear regions were introduced by \citet{pascanu2013number} and \citet{montufar2014number}, to make sense of hierarchical representations in ReLU networks, illustrating how they nonlinearly partition the input space into disjoint cells, and providing lower bounds on density realized by a given architecture.
Several works followed, proving refined bounds on density~\citep{xiong2020number, hanin2019complexity, arora2018understanding, serra2018bounding, raghu2017expressive}. \citet{hanin2019deep} formally study neural networks at initialization, providing average-case bounds over the weight space of MLPs on the local density of linear regions, computed on bounded volumes. Beyond initialization, they hypothesize that the implicit bias of SGD pushes learning towards functions that realize relatively low density -- expressed by a polynomial bound in the number of neurons.

\citet{novak2018sensitivity} estimate sensitivity of dense ReLU networks to input perturbations, as well as density of linear regions, as two independent quantities, along data-driven directions in the input space of trained networks. They observe that while their measure of sensitivity correlates with generalization, density does so only weakly. Similarly, in this work we adopt the same strategy for generating data-dependent directions in the input space, but use them to carry out a systematic investigation of the relationship between density of regions and the nonlinearity of the piece-wise affine functions. Importantly, while \citet{novak2018sensitivity} estimate density by sampling uniformly in the input space, our work accounts for the anisotropic geometry of linear regions, by providing an adaptive algorithm for region discovery, thus improving on the tradeoff between analytical precision and scalability.

\paragraph{Density-agnostic Works on Bounded Variation for Generalizing Networks}

Recent works deals with measures of variation implicitly regularized by optimization, by explicitly accounting for linear regions, but not for their density. \citet{rahaman19spectral}~identify an implicit bias of SGD, by computing the Fourier transform of ReLU networks, and find directions tied with slower spectral decay of ReLU networks. 
While their work directly exploits the piece-wise linearity of ReLU networks, as well as the notion of linear region, their effort is focused on establishing the emergence of the spectral bias.

\citet{lejeune2019implicit} study the complexity of ReLU networks, by introducing a \textit{rugosity} measure based on the tangent Hessian of the function. Their work highlights how commonly used data augmentation implicitly regularizes rugosity. While their work is not directly concerned with the density of linear regions, they identify as meaningful future work the study of variation of piece-wise affine networks, and speculate whether such notion would be more appropriate to study the implicit bias of SGD, as opposed to directly enumerating regions. Our work follows in spirit their observation, and we are able to empirically identify directions in the input space in which uniform density of regions fails to correlate with variation of the underlying function, which itself correlates with learning.

Importantly, existing function variation measures have so far not been studied in the double descent regime, in relation to test error and increased overparameterization. Our work is the first to explicitly tie reduced test error with reduced nonlinearity of ReLU networks.

\paragraph{Empirical Properties of Linear Regions}

\citet{zhang2020empirical} carry out an empirical study of linear regions for small networks trained in practice, relating statistics of linear regions with several neural network hyperparameters. Furthermore, they study how explicit regularization affects similarity of neighbouring linear regions using validation points, and observe that neighbouring regions equally contribute to the prediction of the network. Their study focuses on the effect of different explicit regularization techniques on the statistics of linear regions, while we relate function variation with realized density of linear regions.

\citet{trimmel2021tropex} provide an algorithm for compressing a trained network by using only non-empty linear regions containing training data, which is able to partly recover the performance of the uncompressed network. The authors argue for their extraction method to be better suited at capturing nonlinearity underlying piece-wise affine functions in place of density of linear regions, but do not investigate where density fails in practice. Our work is complementary to theirs in research question and methodology, as we systematically assess the fragility of linear region density as an estimator of nonlinearity for trained networks.

\section{CONCLUSIONS}
\label{sec:conclusions}
    Our experimental investigation, contrasting density of regions to a principled measure of nonlinearity, shows that not all linear regions equally contribute to complexity underlying learning in ReLU networks. Crucially, following model-wise double descent, linear regions increase in local similarity, with corresponding reduction of both nonlinearity and harmful overfitting. Existing density estimates, that equally weigh different linear regions, provide a fragile measure in practice for modern overparameterized networks. While density still serves as an important tool to describe theoretical expressivity of ReLU network architectures, our work supports with empirical evidence the criticism of density as a complexity measure for trained networks. Importantly, our work highlights that estimates of locally linear behaviour for ReLU networks can correlate with complexity and learning. Thus, implicit bias of SGD should be sought in bounded variation of the learned function, rather than by bounding uniform density of regions. We leave to future work the question of further biasing region-counting in the input space, so that redundant regions are weighted less, potentially resulting in a more robust notion of density.

\subsubsection*{Acknowledgments}
This work was partially supported by the Wallenberg AI, Autonomous Systems and Software Program (WASP) funded by the Knut and Alice Wallenberg Foundation, as well as partially funded by Swedish Research Council project 2017-04609. Large-scale experiments were enabled by resources provided by the Swedish National Infrastructure for Computing (SNIC) at Chalmers Centre for Computational Science and Engineering (C3SE) partially funded by the Swedish Research Council through grant agreement no. 2018-05973, as well as by resources provided by the National Supercomputer Centre (NSC), funded by Link\"{o}ping University.

\bibliography{ms}
\bibliographystyle{sty/natbib}

\clearpage
\appendix

\thispagestyle{empty}

\onecolumn \makesupplementtitle

The following presents a detailed description of our methodology and experimental setting. Section~\ref{sec:appendix:setup} describes our experimental setup, with training hyperparameters and hardware used. Section~\ref{sec:appendix:models} describes the network architectures used. Pseudocode for our linear region discovery algorithm is presented in section~\ref{sec:appendix:algorithm}, while section~\ref{sec:appendix:abs_deviation} provides a full derivation of absolute deviation. Then, section~\ref{sec:appendix:paths} details the transformations used to generate data-driven trajectories in the input space. Finally, section~\ref{sec:appendix:experiments} presents additional experiments.

\section{Training Hyperparameters}
\label{sec:appendix:setup}

With the exception of Figure~\ref{fig:exps:double-descent}, all networks are trained using SGD with momentum $0.9$ and batch size $128$, on the CIFAR-10 and CIFAR-100 datasets. Pixel values are normalized using per-channel mean and standard deviation, computed on the full training split of the corresponding dataset. For each dataset, training hyperparameters are selected on a fixed validation split of size $1000$, randomly sampled from the training set. All networks are trained without any explicit regularization enabled (dropout, batch normalization, weight decay), save where specified. For all training settings we run $3$ random seeds, controlling random weight initialization and the ordering of samples during training.

\paragraph{Vanilla and Augmentation Training Settings} To ensure a fair computational budget across different network architectures, convergence criteria are set on CIFAR-10 and CIFAR-100 respectively so that all networks are trained until the training cross entropy loss falls below $0.19$, and $0.25$. For all networks, an initial learning rate of $0.1$ is decayed by a factor of $0.2$ every $150$ epochs. When data augmentation is enabled, training images are randomly shifted by $4$ pixels and randomly flipped horizontally. To avoid overfitting, all networks are trained with weight decay $5\rm{e}-4$.

\paragraph{Overfitting Noisy Labels}
On noisy labels, all networks are trained until $100\%$ training accuracy is reached, following the experimental setup of \citet{zhang2018understanding}. Namely, an initial learning rate of $0.1$ is decayed with a multiplicative factor of $0.95$ at every epoch. For this setting, no data augmentation is used.

\paragraph{Model-Wise Deep Double Descent}
Following the experimental setting of \citet{nakkiran2019deep}, we train ResNets18 of increasing base width $w \in \{1, 2, 4, 8, 16, 24, 32, 64\}$, according to the ResNetv1 architecture~\citep{he2015delving}, consisting of residual blocks grouped in $4$ stages, with respective width ${[}w, 2w, 4w, 8w{]}$. All networks are trained for $400$ epochs with Adam with base learning rate $1\mathrm{e}-4$ on CIFAR-10 with $20\%$ corrupted training labels. We limit the number of training epochs to $400$ to avoid incurring in epoch-wise double descent, and focus on model-wise descent only. All networks are trained with no explicit regularization (especially without batch normalization), nor data augmentation. To ensure stable training, FixUp initialization is used~\citep{zhang2018residual}, without the affine rescaling transformations to avoid altering the ResNet architecture.

\paragraph{Hardware Infrastructure} Our experiments on VGG8 are performed using $8$ NVIDIA V100s with $32$GB of memory, while for ResNet18, linear region counting is performed using $8$ A30s, each with $40$GB of memory.

\section{Network Architectures}
\label{sec:appendix:models}

\begin{table}[]
\caption{Network architectures used in our experiments. Following the notation of \citet{simonyan15very}, Conv$3$-$64$ denotes a convolutional layer of kernel size $3 \times 3$ learning $64$ feature maps. Strides greater than $1$ are denoted by //$s$, e.g.\ //$2$ for stride $2$. For both architectures, each layer save for the last linear one is followed by a ReLU activation. For ResNet18, ``downsample'' convolutions denote residual connections with stride larger than $1$. All trained layers learn bias parameters, save for the $1\times 1$ convolutions in the ResNet downsample blocks.}
\label{tab:appendix:architectures}
\begin{center}
\begin{tabular}{@{}ccc@{}}
\toprule
 &
  VGG8 &
  ResNet18 \\ \midrule
Input size &
  $32 \times 32 \times 3$ &
  $32 \times 32 \times 3$ \\ \midrule
 &
  \begin{tabular}[c]{@{}c@{}}Conv3-6\\ Conv3-6\\ AveragePool (2,2)\end{tabular} &
  Conv3-16 \\ \midrule
Input size &
  $16 \times 16 \times 6$ &
  $32 \times 32 \times 16$ \\ \midrule
 &
  \begin{tabular}[c]{@{}c@{}}Conv3-16\\ Conv3-16\\ AveragePool (2,2)\end{tabular} &
  \begin{tabular}[c]{@{}c@{}}Conv3-16\\ Conv3-16\\ \\ Conv3-16\\ Conv3-16\end{tabular} \\ \midrule
Input size &
  $8 \times 8 \times 16$ &
  $32 \times 32 \times 16$ \\ \midrule
 &
  \begin{tabular}[c]{@{}c@{}}Conv3-64\\ Conv3-64\\ AveragePool (2,2)\end{tabular} &
  \begin{tabular}[c]{@{}c@{}}Conv3-32 //2\\ Conv3-32\\ Downsample: Conv1-32 //2\\ \\ Conv3-32\\ Conv3-32\end{tabular} \\ \midrule
Input size &
  $4 \times 4 \times 64$ &
  $16 \times 16 \times 32$ \\ \midrule
 &
   &
  \begin{tabular}[c]{@{}c@{}}Conv3-64 //2\\ Conv3-64\\ Downsample: Conv1-64 //2\\ \\ Conv3-64\\ Conv3-64\end{tabular} \\ \midrule
Input size &
   &
  $8 \times 8 \times 64$ \\ \midrule
 &
   &
  \begin{tabular}[c]{@{}c@{}}Conv3-128 //2\\ Conv3-128\\ Downsample: Conv1-128 //2\\ \\ Conv3-128\\ Conv3-128\end{tabular} \\ \midrule
Input size &
  $4 \times 4 \times 64$ &
  $4 \times 4 \times 128$ \\ \midrule
 &
  \multicolumn{2}{c}{Global Average Pooling (4, 4)} \\ \midrule
Input size &
  128 &
  128 \\ \midrule
 &
  fc-1024, fc-120 &
  fc-128 \\ \midrule
Depth &
  8 &
  18 \\
Number of parameters &
  \begin{tabular}[c]{@{}c@{}}174116 (CIFAR-10)\\ 185006 (CIFAR-100)\end{tabular} &
  700042 (CIFAR-10) \\ \bottomrule
\end{tabular}
\end{center}
\end{table}

Table~\ref{tab:appendix:architectures} lists all network architectures used. VGG is initialized following the He initialization scheme~\citep{he2015delving}, with bias parameters initialized to zero. ResNet is instead initialized using a modified version of FixUp~\citep{zhang2018residual}, where in place of initializing some layers to zero, each component is sampled i.i.d.\ from a Gaussian distribution of mean zero and standard deviation $1\rm{e}-6$. To preserve piece-wise linearity of the functions parameterized by the networks, average pooling was used in place of max pooling. We note that the goal of the paper is contrasting learned piece-wise affine functions, rather than optimizing networks for extreme performance.

\section{Linear Region Discovery Algorithm}
\label{sec:appendix:algorithm}

This section presents the pseudocode for the linear region discovery algorithm introduced in section~\ref{sec:method:discovery}, together with a complexity analysis.

\begin{algorithm}[t]
    \caption{Find minimum displacement $\lambda$ to cross into the next linear region.}
    \label{alg:appendix:lambda}
    \begin{algorithmic}[1]
    
    \Function{Find-Lambda}{$\mathbf{f}, \mathbf{x}, \mathbf{d}$}
        \State  $\lambda \gets \infty$
        \For{$\ell = 1, \ldots, L$} \Comment{Iterate sequentially through all layers.}
            \State lambdas $\gets \emptyset$ \Comment{Set of candidate lambdas for layer $\ell$.}
            \State lambdas $\gets \frac{-\mathbf{b}^\ell - \mathbf{W}^\ell\mathbf{x}}{\mathbf{W}^\ell\mathbf{d}}$ \Comment{Element-wise division, producing $d_\ell$ candidate lambdas.}
            \State lambdas${[}$lambdas $< \tau{]}\gets \infty$ \Comment{Filter out numerically unstable lambdas.}
            \If{$\min\text{lambdas} < \lambda$}
                \State $\lambda \gets \min \text{lambdas}$
            \EndIf
             \State $\mathbf{x} \gets \mathbf{W}_\epsilon^\ell \mathbf{x} + \mathbf{b}_\epsilon^\ell$ \Comment{Forward pass through layer $\ell$.}
             \State $\mathbf{d} \gets \mathbf{W}_\epsilon^\ell\mathbf{d}$
        \EndFor
        \State\Return $\lambda$
    \EndFunction
    \end{algorithmic}
\end{algorithm}

\paragraph{Linear Region Discovery on Compact 1-Dimensional Domains}  Let $\mathbf{x}_0, \mathbf{x}_1 \in \mathbb{R}^d$ be two distinct points in the input space of a ReLU network $\mathbf{f}: \mathbb{R}^d \to \mathbb{R}^K$, and let $\mathbf{d} = \mathbf{x}_1 - \mathbf{x}_0$ be the corresponding direction vector. Consider the line segment $\bm{\pi}(t): \mathcal{I} = {[}0, 1{]} \to \mathbb{R}^d$ such that $t \mapsto \mathbf{x}_0 + t\mathbf{d}$. As observed in section~\ref{sec:method:discovery}, $\mathbf{f}$ implicitly induces a partition $\mathcal{P}$ of $\bm{\pi} \in \mathbb{R}^d$ into disjoint linear regions, whose entry and exit points along $\bm{\pi}$ in turn induce a partition $\mathcal{P}_{\mathcal{I}}$ of $\mathcal{I}$, with $\mathcal{P}_\mathcal{I} = \{t_\epsilon \in \mathcal{I} : 0 = t_0 < \ldots < t_D = 1\}$. Algorithm~\ref{alg:appendix:algorithm} describes a procedure for numerically determining $\mathcal{P}_\mathcal{I}$.

Starting from $\mathbf{x} = \mathbf{x}_0 \in A_0$, our method travels along $\mathbf{d}$ by computing the smallest displacement $\lambda_0$ to cross a linear region boundary of $A_0$ along $\mathbf{d}$. Afterward, $\mathbf{x}$ is moved to the linear region boundary, $\mathbf{x} = \mathbf{x} + \lambda_0 \mathbf{d}$, and the procedure is repeated until $\mathbf{x}$ falls in the same linear region as $\mathbf{x}_1$.

For each linear region discovered, the smallest lambda to cross one of its boundaries is computed by solving Equation~\ref{eq:method:lambda}, as detailed in Algorithm~\ref{alg:appendix:lambda}. For numerical stability, when solving for $\lambda$, all values below the sensitivity threshold $\tau$ are discarded.

Linear region membership of $\mathbf{x} \in A_\epsilon$ is defined by using the corresponding activation pattern $\mathbf{x}$ $\big{[}\vect(\mathbf{M}^{1}_{\mathbf{x}}), \ldots, \vect(\mathbf{M}^{L}_{\mathbf{x}}) \big{]}$, obtained when forward-passing $\mathbf{x}$ through $\mathbf{f}$.

Finally, the algorithm terminates under any of the following conditions. If the activation pattern of $\mathbf{x}$ is the same as the one of $\mathbf{x}_1$, then all linear regions have been discovered and the procedure completes. Furthermore, to ensure numerical stability, Algorithm~\ref{alg:appendix:algorithm} terminates also if no finite $\lambda$ can be found, or if $\lambda$ overshoots $\mathbf{x}_1$. The former condition occurs only when --- either exactly or approximately --- $\mathbf{d}^\ell$ is contained in all linear region boundaries, for every neuron in the network. The latter occurs when, for every neuron in the network, $\mathbf{d}$ is approximately parallel to each hyperplane. We didn't observe any such case in practice.
  
In section~\ref{sec:experiments}, we run our algorithm on data-driven trajectories obtained by connecting multiple line-paths $\bm{\pi}$ to form a closed loop in the input space. Throughout our experiments, computations are performed with double precision, with sensitivity $\tau = 1\rm{e}-6$, for normalized pixel values and unnormalized direction $\mathbf{d}$\footnote{In the scale of normalized directions $\hat{\mathbf{d}} = \frac{\mathbf{x}_1 - \mathbf{x}_0}{\| \mathbf{x}_1 - \mathbf{x}_0 \|_2}$, $\tau$ is approximately $1\rm{e}-9$.}.

\paragraph{Complexity Analysis} A na\"{i}ve implementation of algorithms~\ref{alg:appendix:lambda} and~\ref{alg:appendix:algorithm} entails multiple sequential bottlenecks. For a set of $N$ paths $\{\bm{\pi}_n\}_{n=1}^N$, each consisting of $A$ line segments, estimating the respective density $D_n$ along each $\bm{\pi}_n$, involves iterating through each path, line segment, layer $\ell$ of $\mathbf{f}$, as well as neuron in each layer, for a total run-time complexity of $\mathcal{O}(2 \cdot N A L D V)$, with $D := \max_{n} D_n$, $V = \sum\limits_{\ell = 1}^L d_\ell$, and the factor of $2$ arising from the need of forward-propagating both $\mathbf{x}$ and $\mathbf{d}$ in Algorithm~\ref{alg:appendix:algorithm}. Specifically, such complexity is the same as for analytical methods, as described by~\citet{zhang2020empirical, hanin2019deep}.

In our implementation, we collate all line segments into a single set of $N\cdot A$ line paths, and estimate density in parallel over batches of $B = 1024$ lines at a time, reducing the data-dependent complexity to $\mathcal{O}(\frac{N A}{B})$. Furthermore, for each layer, Algorithm~\ref{alg:appendix:lambda} allows to parallelize computation across all neurons composing the layer, bringing down the complexity to $\mathcal{O}(2\frac{N A}{B}LD)$. Additionally, in our implementation, we batch $\mathbf{x}$ and $\mathbf{d}$ together, so that each $\lambda$ can be estimated by running a single forward pass~\footnote{In our code this is achieved by exploiting Pytorch's forward-hooks.}, bringing down the complexity to $\mathcal{O}(\frac{N A}{B}LD)$. Finally, we note that sampling-based methods would require estimating $D$ a priori, making counting either imprecise or more expensive than needed. Our algorithm exactly recovers all linear regions.

We observe that the dependency on the depth $L$ of $\mathbf{f}$ is intrinsic in the nature of forward passes in feed-forward networks, and cannot currently be removed.

In conclusion, to quantify complexity in practice, we note that the run-time complexity of training the same network $\mathbf{f}$ on a training set of size $S$, using SGD with batch size $B$, for $E$ epochs, involves running $\big\lfloor\frac{S}{B}\big\rfloor E$ forward passes, with total\footnote{This is clearly a lower bound on the complexity of training, since backward passes are not accounted for. Furthermore, as is the case for training, optimizations like distributed data-parallel paradigms can be applied to our method, allowing to process larger batch sizes, thus reducing the number of forward passes required to $\mathcal{O}(LD)$.} complexity $\mathcal{O}(\frac{S}{B}EL)$. In practice, for a network trained on the CIFAR-10 training split, of size $S = 50000$ samples, for $E = 200$ epochs, the number of forward passes is approximately $78000$. For VGG8 such number is thus similar to the average observed density $D$ (c.f.\ Figure~\ref{fig:exps:task-ranking}), while it is $4$ times lower for ResNet18.

\begin{algorithm}
    \caption{Linear region discovery on compact 1-D domains.}
    \label{alg:appendix:algorithm}
    \begin{algorithmic}[1]
    
    \Function{FindLinearRegions}{$\mathbf{f}, \mathbf{x}_0, \mathbf{x}_1, \tau$}
        \State $\mathbf{c} \gets$ \Call{ActivationPattern}{$\mathbf{f}, \mathbf{x}_0$}
        \State $\mathbf{c}_1 \gets$ \Call{ActivationPattern}{$\mathbf{f}, \mathbf{x}_1$}
        \State $\mathbf{d} \gets \frac{\mathbf{x}_1 - \mathbf{x}_0}{\|\mathbf{x}_1 - \mathbf{x}_0\|_2}$
        \State $\mathbf{x} \gets \mathbf{x}_0$
        \State $\mathcal{P}_\mathcal{I} \gets \emptyset$
        \While{$\mathbf{c} \ne \mathbf{c}_1$}
            \State $\lambda \gets$ \Call{FindLambda}{$\mathbf{f}, \mathbf{x}, \mathbf{d}, \tau$}
            \State $\mathcal{P}_\mathcal{I} \gets \mathcal{P}_\mathcal{I} \cup \{ \lambda\}$
            \If{$\lambda = \infty$} \Comment{Degenerate case. No finite lambda found.}
                \State \textbf{break}
            \EndIf
            \State $\mathbf{x} \gets \mathbf{x} + \lambda \mathbf{d}$ \Comment{Step until linear region boundary}
            \If{$\|\mathbf{x}_0 - \mathbf{x}_1 \|_2 < \|\mathbf{x} - \mathbf{x}_0 \|_2$} \Comment{Overshot $\mathbf{x}_1$}
                \State \textbf{break}
            \EndIf
            \State $\mathbf{c} \gets$ \Call{ActivationPattern}{$\mathbf{f}, \mathbf{x}$}
        \EndWhile
        \State\Return $|\mathcal{P}_\mathcal{I}|, \mathcal{P}_\mathcal{I}$ \Comment{Return density, as well as all boundary points.}
    \EndFunction
    
    \end{algorithmic}
\end{algorithm}

\section{Computing Absolute Deviation}
\label{sec:appendix:abs_deviation}

In this section, we present the full derivation of absolute deviation, introduced in section~\ref{sec:method:deviation}. Recalling equation~\ref{eq:method:deviation},
\begin{equation*}
\begin{aligned}
  \int_{\bm{\pi}} | f^k(\mathbf{x}) - a^k(\mathbf{x})| d\mathbf{x} = & \sum\limits_{\epsilon = 0}^{D -1} \int_{t_\epsilon}^{t_{\epsilon +1}} |f^k(\bm{\pi}(t)) - a^k(\bm{\pi}(t)) | \cdot \|\dot{\bm{\pi}}(t) \| dt \\
 ~ = & \sum\limits_{\epsilon = 0}^{D -1} \int_{t_\epsilon}^{t_{\epsilon +1}} |f^k(\mathbf{x}_0 + t\mathbf{d}) - a^k(\mathbf{x}_0 + t\mathbf{d}) | \cdot \|\mathbf{d} \| dt \\
 ~ = & \|\mathbf{d} \| \cdot \sum\limits_{\epsilon = 0}^{D -1} \int_{t_\epsilon}^{t_{\epsilon +1}} \big(|f_\epsilon^k(\mathbf{x}_0) - f_0^k(\mathbf{x}_0) + t \big( f_\epsilon^k(\mathbf{x}_1) - f_\epsilon^k(\mathbf{x}_0) - f_1^k(\mathbf{x}_1) + f_0^k(\mathbf{x}_0) \big)|\big) dt 
\end{aligned}
\end{equation*}
we note that, on each linear region $A_\epsilon$, the affine function inside the absolute value of the integrand is not necessarily monotonic, as $f^k$ and $a^k$ may intersect at 
\begin{equation}
    Z = \frac{f_0^k(\mathbf{x}_0) - f_\epsilon^k(\mathbf{x}_0)}{f_\epsilon^k(\mathbf{x}_1) - f_\epsilon^k(\mathbf{x}_0) -  f_1^k(\mathbf{x}_1) + f_0^k(\mathbf{x}_0)}
    \label{eq:appendix:intersection}
\end{equation}

If this is the case, then the integral over $A_\epsilon$ can be split as the sum of integrals over two sub-intervals ${[}t_\epsilon, t_*{]} \cup {[}t_*, t_{\epsilon +1}{]}$, yielding:
\begin{equation}
\begin{aligned}
  \int_{\bm{\pi}} | f^k(\mathbf{x}) - a^k(\mathbf{x})| d\mathbf{x} = & \|\mathbf{d} \| \cdot \sum\limits_{\epsilon = 0}^{D -1} \int_{t_\epsilon}^{t_*} |f_\epsilon^k(\mathbf{x}_0) - f_0^k(\mathbf{x}_0) + t \big( f_\epsilon^k(\mathbf{x}_1) - f_\epsilon^k(\mathbf{x}_0) - f_1^k(\mathbf{x}_1) + f_0^k(\mathbf{x}_0) \big)| dt + \\
  ~ + &  \|\mathbf{d} \| \cdot \sum\limits_{\epsilon = 0}^{D -1} \int_{t_*}^{t_{\epsilon + 1}} |f_\epsilon^k(\mathbf{x}_0) - f_0^k(\mathbf{x}_0) + t \big( f_\epsilon^k(\mathbf{x}_1) - f_\epsilon^k(\mathbf{x}_0) - f_1^k(\mathbf{x}_1) + f_0^k(\mathbf{x}_0) \big)| dt
\end{aligned}
\end{equation}

Otherwise --- if the intersection between $f^k$ and $a^k$ occurs outside of $A_\epsilon$ --- we take $t_*$ as \mbox{$t_* = \max\{t_\epsilon, \min\{Z, t_{\epsilon+1} \}\}$,} which can be efficiently solved as a simple linear programme.

Finally, for each $k = 1, \ldots, K$, absolute deviation is given by
\begin{equation}
    \begin{aligned}
 \int_{\bm{\pi}} | f^k(\mathbf{x}) - a^k(\mathbf{x})| d\mathbf{x} = & \|\mathbf{d}\| \Big((t_* -t_\epsilon)|f_\epsilon^k(\mathbf{x}_0) - f_0^k(\mathbf{x}_0)| +
        (t_{\epsilon+1} - t_*) |f_\epsilon^k(\mathbf{x}_0) - f_0^k(\mathbf{x}_0)| &+ \\
        & \frac{t_*^2 - t_\epsilon^2}{2}|f_\epsilon^k(\mathbf{x}_1) - f_\epsilon^k(\mathbf{x}_0) - f_1^k(\mathbf{x}_1) + f_0^k(\mathbf{x}_0)| &+ \\
        & \frac{t_{\epsilon+1}^2 - t_*^2}{2}|f_\epsilon^k(\mathbf{x}_1) - f_\epsilon^k(\mathbf{x}_0) - f_1^k(\mathbf{x}_1) + f_0^k(\mathbf{x}_0)| \Big)
    \end{aligned}
\end{equation}

\section{Data-Driven Trajectories}
\label{sec:appendix:paths}

Each data-driven trajectory $\bm{\pi}_n$ in section~\ref{sec:experiments} is obtained from a starting sample $\mathbf{x}_n^0$ and $A$ augmented versions $\{\mathbf{x}_n^a : a = 1, \ldots, A -1 \}$, so that $\bm{\pi}_n$ is composed of line paths connecting each $\mathbf{x}_n^a$ with $\mathbf{x}_n^{a +1 \% A}$, for $a = 0, \ldots, A -1$, forming a closed loop. 

Each augmented image $\mathbf{x}_n^a$ is obtained by translating the starting point $\mathbf{x}_n^0$ along a circular trajectory of radius $r = 4$, with translation parameter $s = ( r \cos \alpha_a, r \sin \alpha_a)$, for $\alpha_a = a \cdot \frac{2\pi}{A}$. Specifically, $\mathbf{x}_n^0$ is first padded by $2$ pixels on each edge, then translated along the circular trajectory, and finally cropped back to its original resolution. Finally, in addition to ~\citet{novak2018sensitivity}, we ensure that no edge artifacts are introduced in the augmented images in the transformation process.

\section{Additional Experiments}
\label{sec:appendix:experiments}

\begin{figure}[t!]
    \begin{center}
        \subfloat[\label{fig:appendix:cdf-compare-vgg}]{\includegraphics[width=0.35\linewidth]{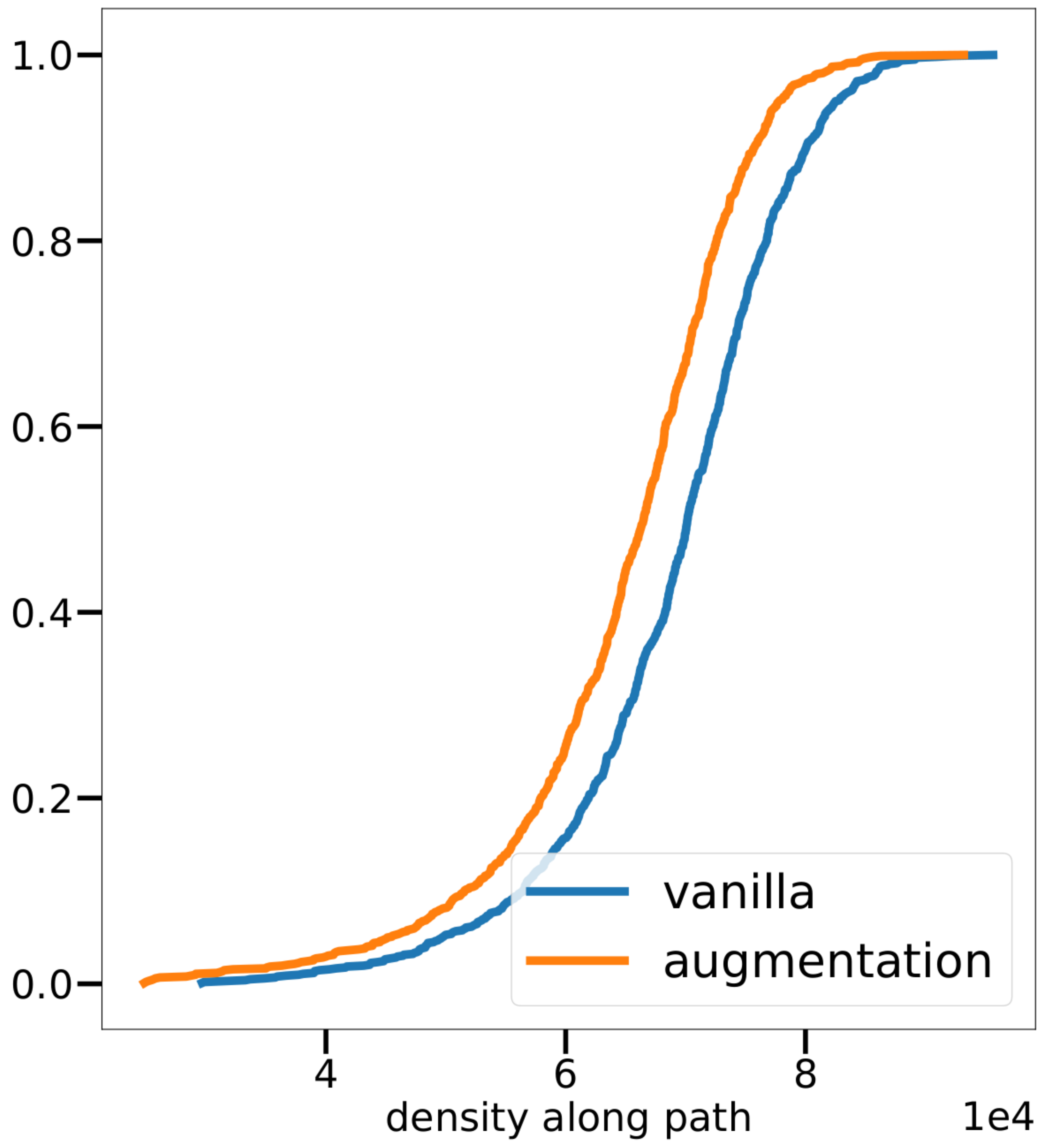}} ~
        \subfloat[\label{fig:appendix:cdf-compare-resnet}]{\includegraphics[width=0.34\linewidth]{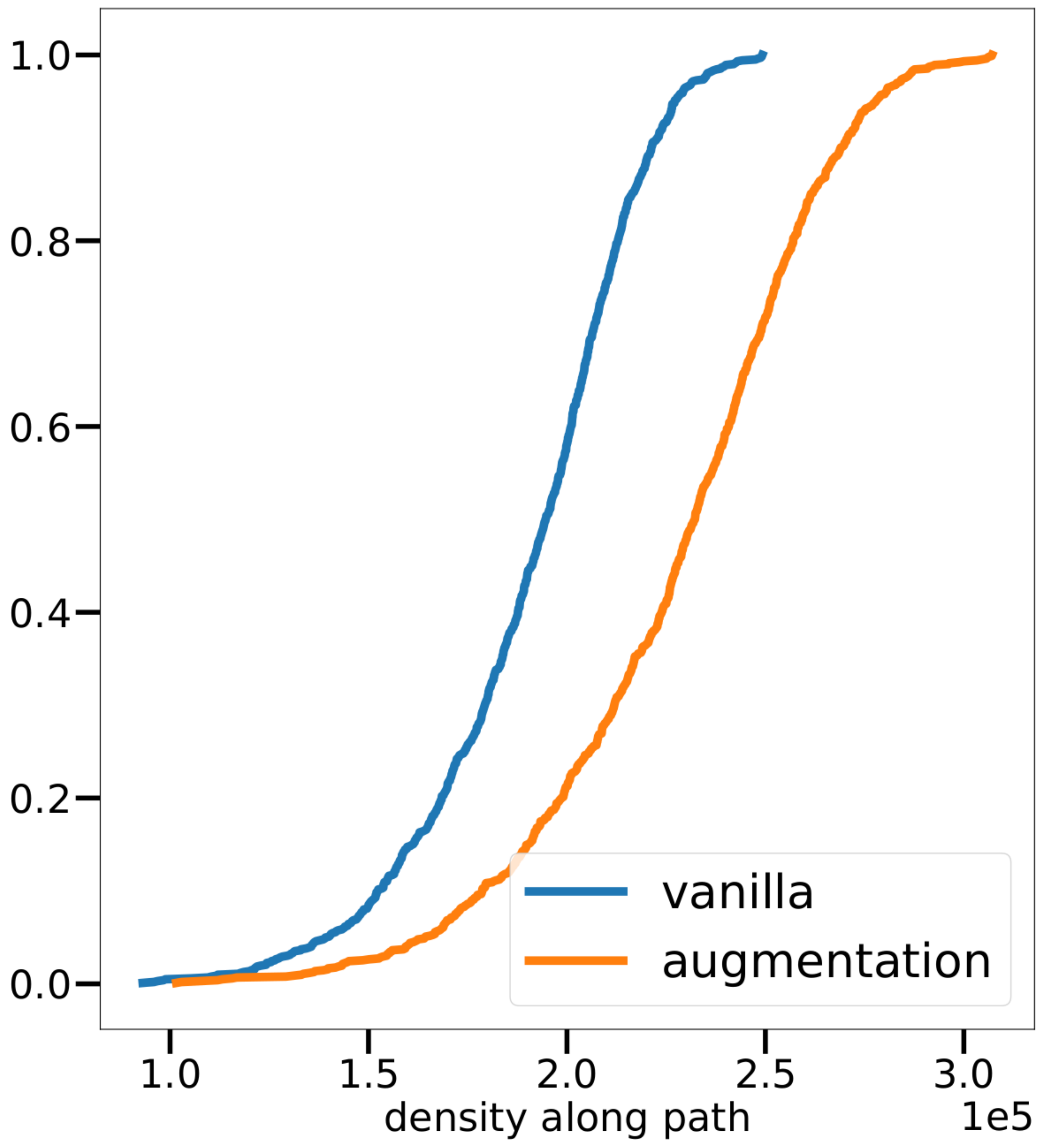}}
    \end{center}
    \caption{ECDF of density along data-driven paths on CIFAR-10 samples, for a network trained with data augmentation and without (``vanilla''). a) VGG8. b) ResNet18.}
    \label{fig:appendix:task-ranking-augmentation}
\end{figure}

\paragraph{Ablating Data-Driven Trajectories}

\begin{figure}[t!]
    \begin{center}
        \includegraphics[width=0.5\linewidth]{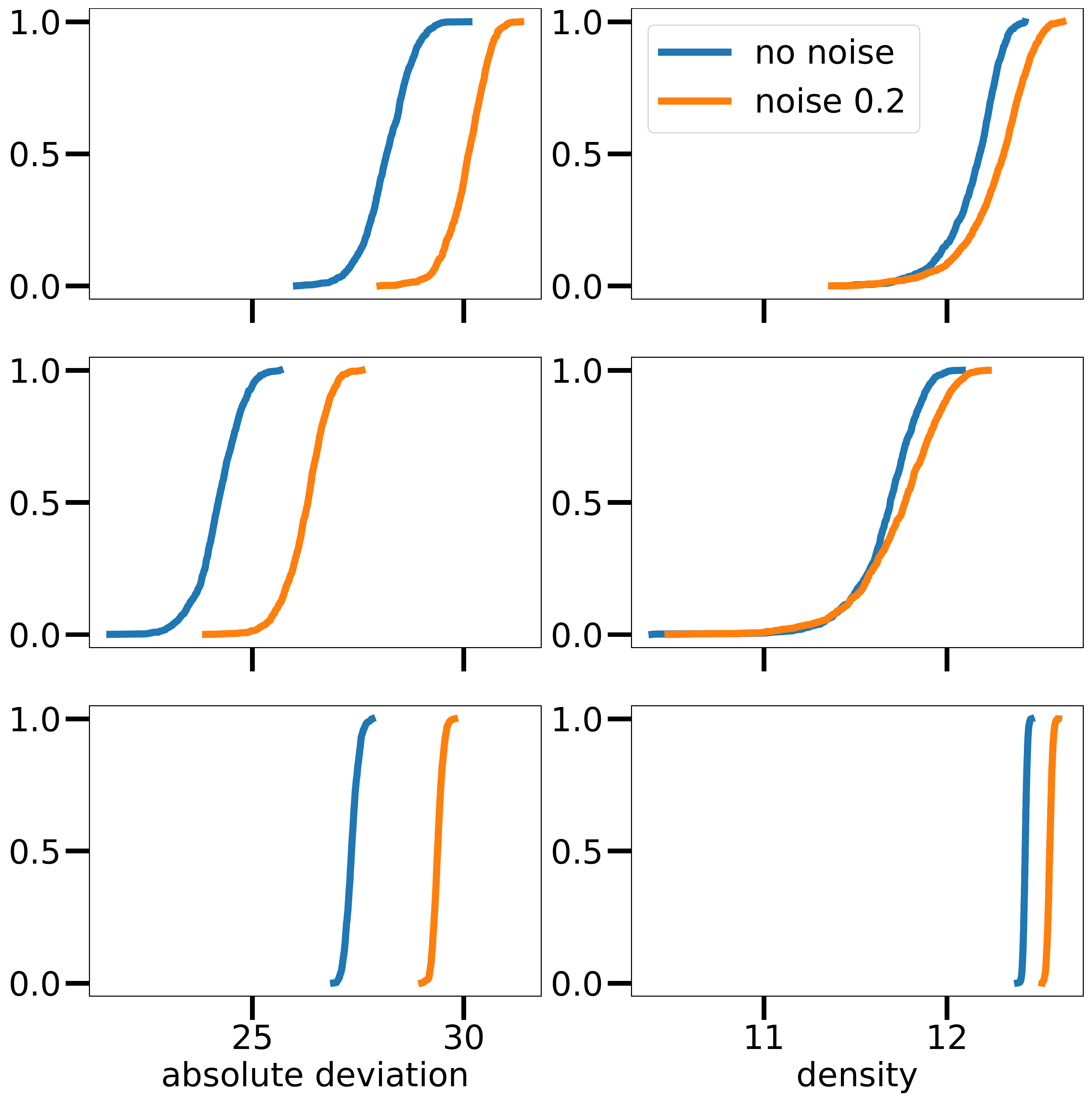}
    \end{center}
    \caption{(Log x-scale) Absolute deviation and density for circular trajectories (top); open paths connecting weak augmentations of a base train image (middle); circular paths connecting uniform noise with per-pixel statistics as the CIFAR-10 training split (bottom).}
   \label{fig:appendix:ablations}
    
\end{figure}

In this section, we consider alternatives to estimating density and absolute deviation along closed loops in the input space. We begin by recalling that, since both density and absolute deviation are computed as line integrals on compact 1D supports, alternative strategies, like MC-based sampling of random directions in pixel space, increase chances of double counting linear regions, making computations more memory intensive. Hence, 1-D paths are a core contribution \wrt scalability of our method. Figure~\ref{fig:appendix:ablations} presents ablations on circular paths, in which three alternatives are considered, namely closed loops as in the main experiments (top), open paths connecting weak augmentations ($1$-pixel shifts) of a base training image (middle), and finally closed paths connecting uniform random noise with the same per-pixel statistics as the CIFAR-10 training split (bottom).

Under all settings, absolute deviation correctly separates ResNet18s respectively trained on clean as well as $20\%$ noisy labels on CIFAR-10, while density struggles on trajectories that lie close to the train data. Finally, both density and deviation are almost uniformly distributed if measured away from the training data, corroborating in the setting of modern networks what observed by \citet{novak2018sensitivity}.

\paragraph{Distribution of Density for Networks Trained with Data Augmentation} Extending Figure~\ref{fig:exps:task-ranking}, we study the distribution of density values for VGG8 and ResNet18 trained with and without data augmentation, reporting our findings in Figure~\ref{fig:appendix:task-ranking-augmentation}. It can be observed that, while density correctly separates VGG8 trained with and without data augmentation, the distance between the two cumulative distribution functions is less marked, in contrast with absolute deviation in Figure~\ref{fig:exps:dist-population}, which more strongly distinguishes between smoothness induced by data augmentation, and vanilla training. Finally, the ECDF curves are ranked in the opposite order for ResNet18 (Figure~\ref{fig:appendix:task-ranking-augmentation}, right), which produces more regions when data augmentation is enabled. This shows that, unike absolute deviation,  density is an unreliable predictor of nonlinearity for trained networks, producing inconsistent results when different architectures are compared.

\begin{figure}[b!]
    \begin{center}
        \subfloat[]{\includegraphics[width=0.33\linewidth]{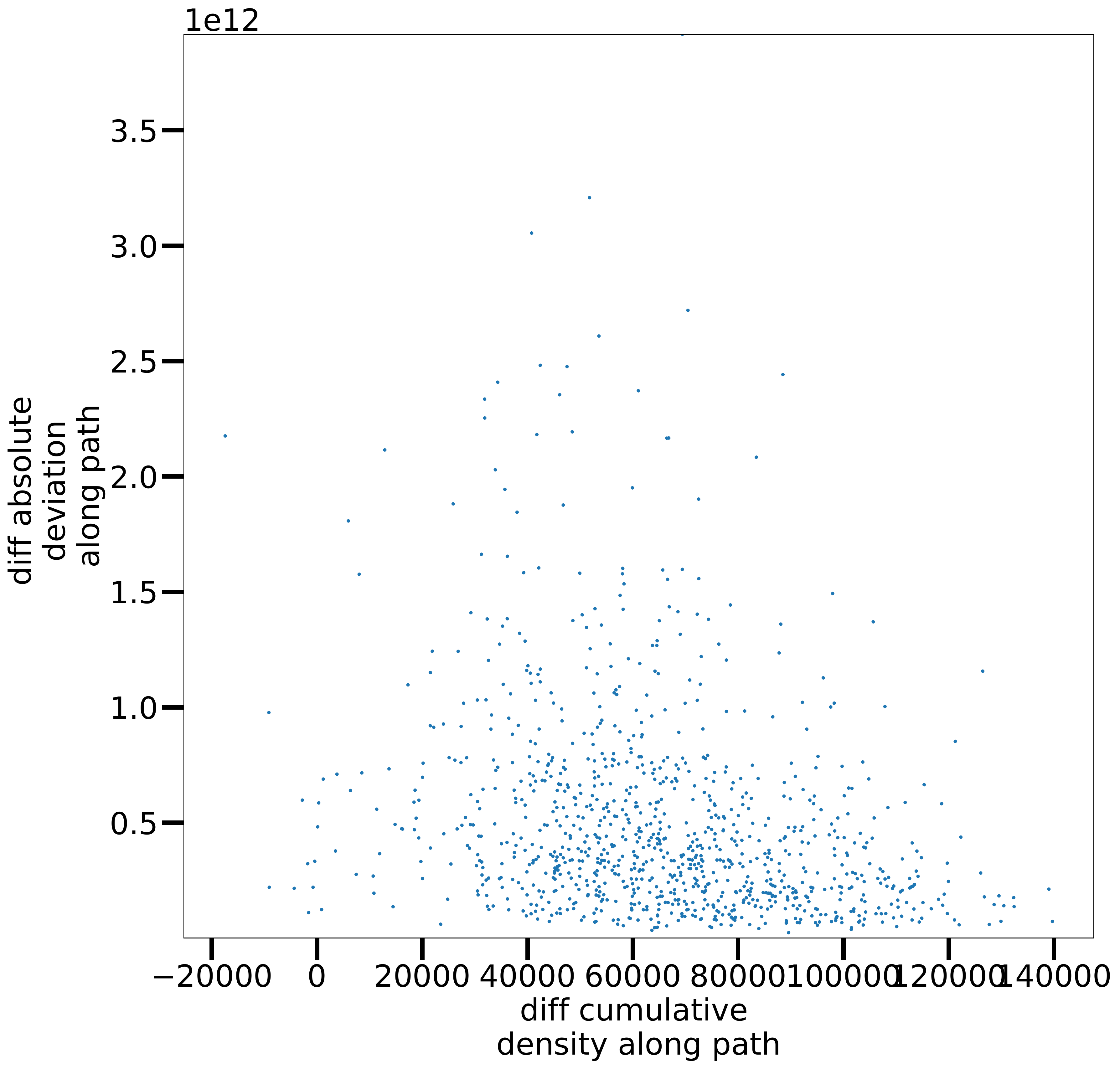}} ~
        \subfloat[]{\includegraphics[width=0.33\linewidth]{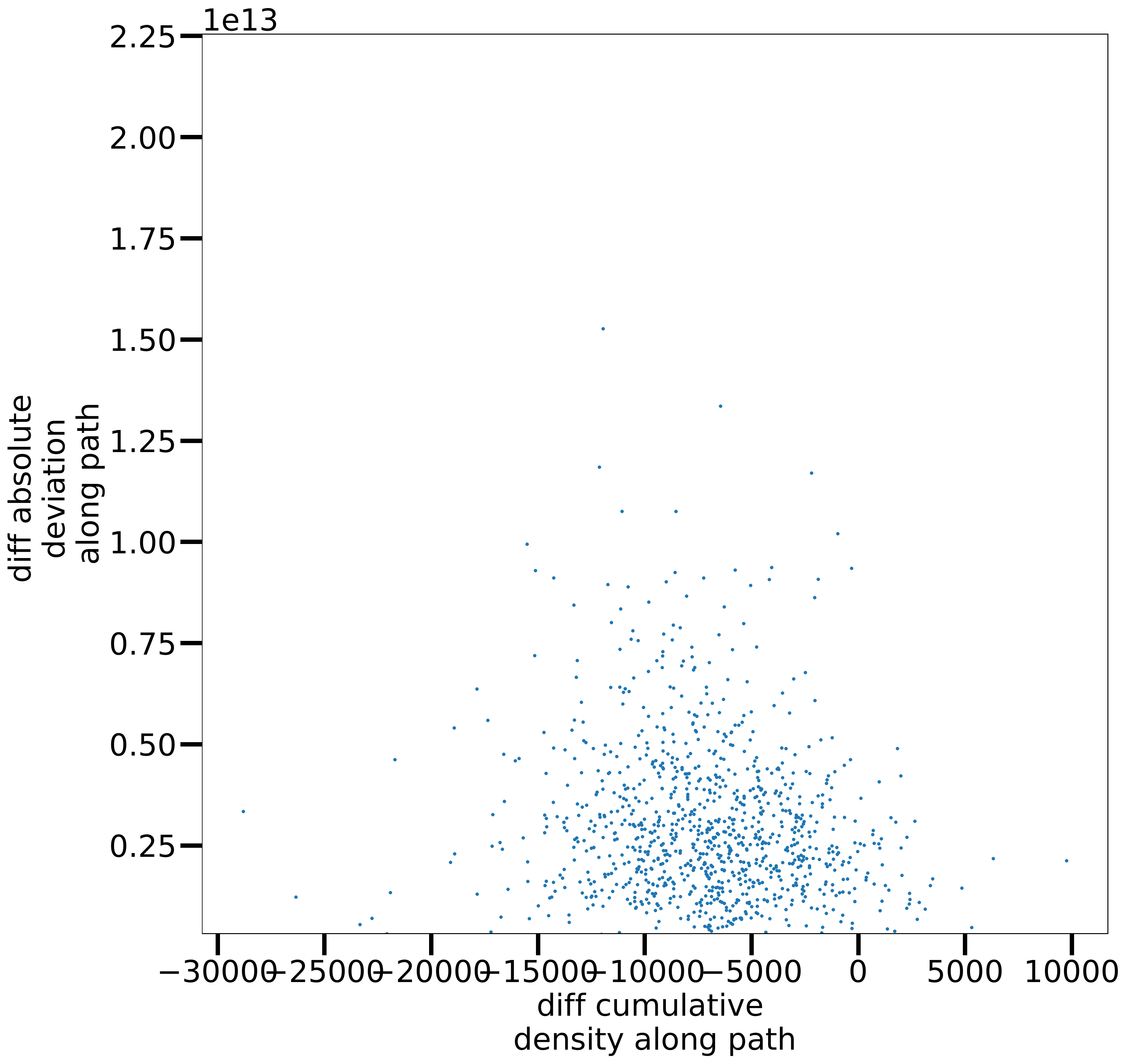}} ~
        \subfloat[]{\includegraphics[width=0.33\linewidth]{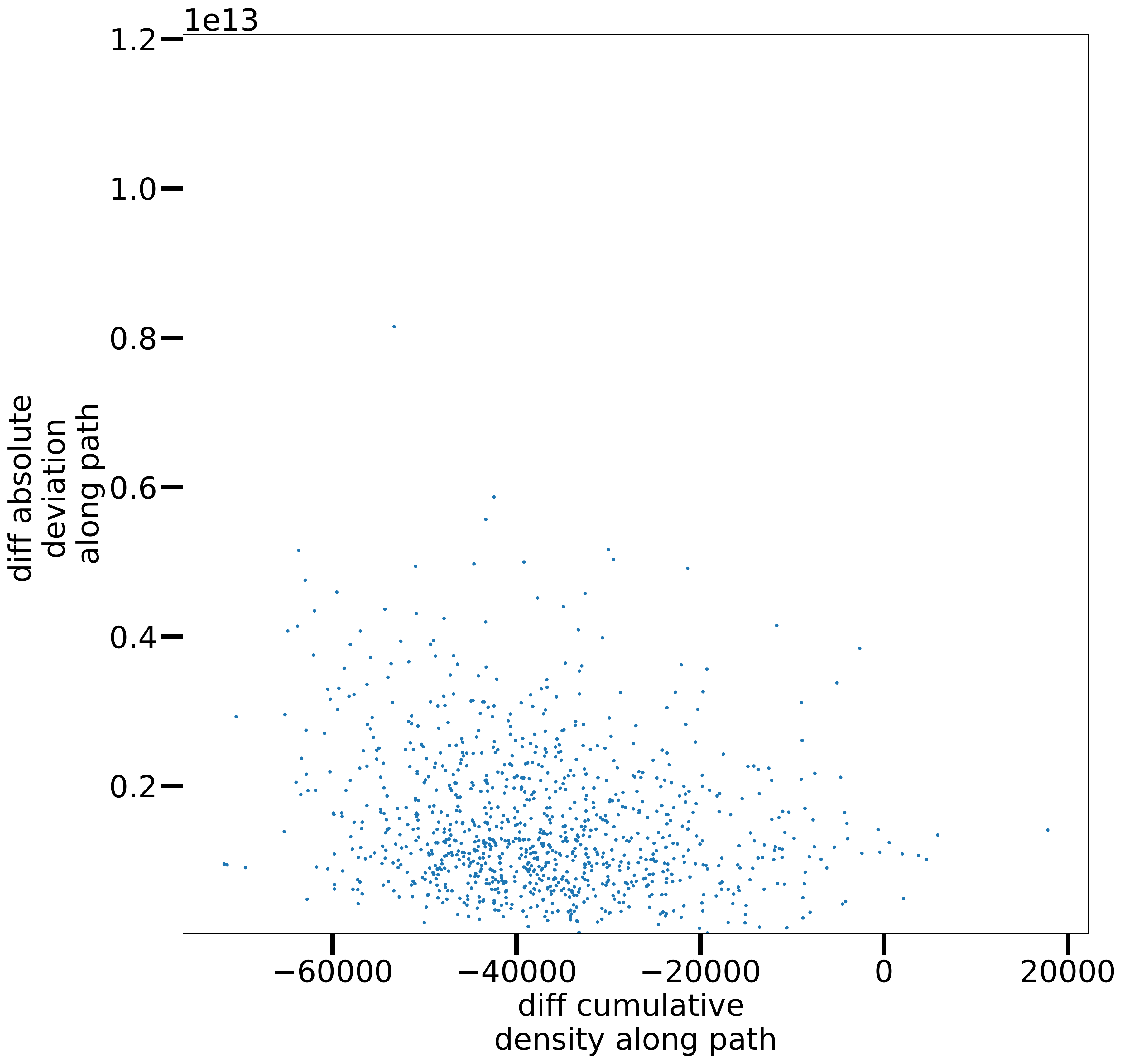}} ~
    \end{center}
    \caption{Distribution of per-path paired differences of density (x-axis) and absolute deviation (y-axis). (Left) Paired differences for a ResNet18 trained with data augmentation on CIFAR-10 vs the same network at initialization. Here, both density and absolute deviation are able to strongly detecting increased nonlinearity at the level of individual paths. (Middle) Paired differences for a VGG8 trained on CIFAR-100 with and without data augmentation. While absolute deviation captures the regularity induced by data augmentation, density becomes a noisy estimator of nonlinearity. (Right) Paired differences for a ResNet18 trained on CIFAR-10 with and without data augmentation. Similarly to VGG8/CIFAR-100, density fails to measure increased nonlinearity when data augmentation is disabled.}
    \label{fig:appendix:differences}
\end{figure}

\paragraph{Distribution of Paired Differences} Complementing Table~\ref{tab:exps:differences}, this section provides a few representative examples of the distribution of instance-level differences, showing how density and absolute deviation change on individual paths when the training settings change. Figure~\ref{fig:appendix:differences} shows the distribution of density and absolute deviation for three settings. First, when comparing a trained ResNet18 on CIFAR-10 with its initialization (left), density can meaningfully capture increased nonlinearity, as showed by the distribution of differences being shifted towards positive values ($0.99 \pm 0.01$ positive differences for density). Second, for a VGG8 trained with and without data augmentation on CIFAR-100 (middle), we see how density becomes a noisy estimator of nonlinearity, as indicated by the distribution of differences being shifted towards negative values ($0.38 \pm 0.24$ positive differences for density). Third, for a ResNet18 trained with and without augmentation on CIFAR-10 (right), density fails to capture the difference in nonlinearity, with the distribution of differences being strongly biased towards negative values ($0.03 \pm 0.03$ positive differences for density). In contrast, absolute deviation consistently and more reliably detects increased nonlinearity at the level of individual paths (as shown by always positive distribution values), thus better capturing the nonlinear behaviour of piece-wise affine functions for ReLU networks.

\end{document}